\documentclass[10pt,twocolumn,letterpaper]{article}

\usepackage[pagenumbers]{wacv}      

\usepackage{xcolor}
\usepackage{pifont}
\newcommand{\cmark}{\textcolor{green!70!black}{\ding{51}}}
\newcommand{\xmark}{\textcolor{red}{\ding{55}}}
\usepackage{multirow}
\usepackage{graphicx}
\usepackage{subcaption} 

\usepackage{array}
\makeatletter
\renewcommand\@fnsymbol[1]{%
  \ifcase#1
    \or $\dagger$%
    \or $\ast$%
    \or $\ddagger$%
    \or $\mathsection$%
    \or $\mathparagraph$%
    \or $\|$%
    \or $\ast\ast$%
    \or $\dagger\dagger$%
    \or $\ddagger\ddagger$%
  \fi}
\makeatother

\definecolor{wacvblue}{rgb}{0.21,0.49,0.74}
\usepackage[pagebackref,breaklinks,colorlinks,allcolors=wacvblue]{hyperref}

\def\wacvPaperID{1683} 
\def\confName{WACV}
\def\confYear{2026}

\title{\vspace{-1cm}SCALEX: \textbf{S}calable \textbf{C}oncept and \textbf{L}atent \textbf{Ex}ploration for Diffusion Models\vspace{-0.5cm}}

\author{E. Zhixuan Zeng, Yuhao Chen\thanks{Corresponding author.} , Alexander Wong\\
University of Waterloo\\
{\tt\small \{ezzeng, yuhao.chen1, alexander.wong\}@uwaterloo.ca}
}

\begin{document}
\maketitle
\begin{abstract}

\vspace{-0.4cm}Image generation models frequently encode social biases, including stereotypes tied to gender, race, and profession. Existing methods for analyzing these biases in diffusion models either focus narrowly on predefined categories or depend on manual interpretation of latent directions. These constraints limit scalability and hinder the discovery of subtle or unanticipated patterns.

We introduce \textbf{SCALEX}, a framework for scalable and automated exploration of diffusion model latent spaces. SCALEX extracts semantically meaningful directions from H-space using only natural language prompts, enabling zero-shot interpretation without retraining or labelling. This allows systematic comparison across arbitrary concepts and large-scale discovery of internal model associations.
We show that SCALEX detects gender bias in profession prompts, ranks semantic alignment across identity descriptors, and reveals clustered conceptual structure without supervision. By linking prompts to latent directions directly, SCALEX makes bias analysis in diffusion models more scalable, interpretable, and extensible than prior approaches.

\end{abstract}
    
\vspace{-0.5cm}\section{Introduction}
\label{sec:intro}

Latent diffusion models \cite{rombachHighResolutionImageSynthesis2022} have become a central tool for image generation in fields such as design, advertising, and media production. Despite their ability to produce high-quality and diverse content, they are also known to encode and reproduce various biases. Stereotypical representations of gender, race, and profession are common, often reflecting imbalances in the underlying training data \cite{wuRevealingGenderBias2025, liT2ISafetyBenchmarkAssessing2025}.

Efforts to evaluate fairness in diffusion models typically rely on predefined categories like gender \cite{wuRevealingGenderBias2025}. T2ISafety \cite{liT2ISafetyBenchmarkAssessing2025} provides a comprehensive benchmark across multiple axes, but still targets only known biases. These methods are useful for confirming expected issues but cannot surface implicit associations. Many subtle patterns go undetected, such as chopsticks consistently appearing with Chinese food, or wine glasses appearing only half full \cite{orgadEditingImplicitAssumptions2023}.

A broader strategy for understanding model behavior is to analyze its latent space. Semantic directions, defined as vector differences corresponding to interpretable changes, have enabled concept manipulation and interpretability in word embeddings \cite{mikolovLinguisticRegularitiesContinuous2013} and GANs \cite{radfordUnsupervised2015}. 

Applying similar techniques to diffusion models may offer insight into how they represent abstract attributes and learned associations. However, existing approaches fail to enable scalable bias analysis due to three main limitations.
First, editing-focused methods like SEGA \cite{brackSEGAInstructingDiffusion2023} and Concept Algebra \cite{wangConceptAlgebraScoreBased2023} produce latent vectors that are specific to initial noise and timestep. These vectors are not reusable and do not reflect stable semantic directions, making them unsuitable for bias analysis.
Second, classifier-based methods \cite{liSelfDiscoveringInterpretableDiffusion2023, pariharBalancingActDistributionGuided2024} operate in H-space \cite{kwonDiffusionModelsAlready2022}, a semantically meaningful representation extracted from the model's middle U-Net layer. This is more consistent across timesteps and captures higher-level semantics. However, they require training a separate direction for each concept, limiting scalability.
Third, methods such as NoiseCLR \cite{dalvaNoiseCLRContrastiveLearning2023} and Park et al. \cite{parkUnsupervisedDiscoverySemantic2023} extract open-ended semantically meaningful directions without labels. But each direction must be manually interpreted, which restricts the scalability of analysis.

To address these limitations, we introduce \textbf{SCALEX}, a framework for scalable bias analysis and concept discovery in diffusion models. SCALEX aligns latent directions with natural language prompts, enabling zero-shot interpretation without manual labeling or concept-specific training.

A key technical challenge is that while trained directions in H-space are semantically meaningful and reusable, the prompt-conditioned activations observed during inference remain entangled and difficult to interpret. SCALEX addresses this by applying Latent Consistency Models (LCMs) \cite{luoLatentConsistencyModels2023}, which stabilize H-space and enable single-step extraction of prompt-aligned directions without additional training or supervision.

Given prompt-aligned latent directions, SCALEX enables scalable and automated analysis of conceptual structure in diffusion models. Because each direction is tied to a natural language prompt, it can be interpreted directly, eliminating the need for manual labeling or classifier training. This allows systematic comparisons across arbitrary concepts, such as measuring gender associations in profession prompts or ranking how closely different identities align with specific attributes. Beyond targeted analyses, SCALEX also supports open-ended discovery through clustering and visualization, revealing emergent groupings without relying on predefined categories.

In summary, \textbf{SCALEX enables scalable analysis of diffusion models by:}
\begin{itemize}
    \item \textbf{Prompt-Aligned Latent Directions,} eliminating manual interpretation.
    \item \textbf{Direct, training-free, single-step vectors} of prompt-aligned semantic directions from H-space using Latent Consistency Models \cite{luoLatentConsistencyModels2023}, eliminating the need for iterative optimization or classifier-based supervision.
    \item \textbf{Scalable evaluation tools} for bias analysis via descriptor ranking, clustering, and concept visualization across arbitrary prompts.
\end{itemize}

\vspace{-3pt}
\section{Related work}
\vspace{-2pt}

\begin{table*}[t]
\centering
\footnotesize
\renewcommand{\arraystretch}{1.3}
\begin{tabular}{>{\raggedright}p{1.8cm} >{\raggedright}p{1.3cm} >{\raggedright}p{3.2cm} >{\raggedright}p{1.2cm} >{\raggedright}p{1.8cm} >{\raggedright}p{2.1cm} p{3cm}}
\textbf{Method} & \textbf{Latent Space} & \textbf{Discovery Method} & \textbf{Reusable Vectors} & \textbf{No Training Required} & \textbf{No Manual \hspace{1cm} Interpretation} & \textbf{Use Case} \\[2pt]

\hline

\textbf{SEGA}\cite{brackSEGAInstructingDiffusion2023} & Score space & Gradient-based guidance vectors from prompt & \xmark~No & \cmark~Yes & \cmark~Yes (prompt specifies concept) & Image editing \\
\hline

\textbf{Concept \hspace{2cm} Algebra} \cite{wangConceptAlgebraScoreBased2023} & Latent noise space & Projects noisy latents onto concept subspaces & \xmark~No & \cmark~Yes  & \cmark~Yes (prompt specifies concept) & Image editing \\
\hline

\textbf{NoiseCLR} \cite{dalvaNoiseCLRContrastiveLearning2023} & Latent noise space & Learns concept vectors via contrastive learning & \cmark~Yes & \xmark~No & \xmark~No & Latent direction discovery \\
\hline

\textbf{Park et al. \cite{parkUnsupervisedDiscoverySemantic2023}} & H-space & Jacobian between the latent space and feature space & \cmark~Yes & \cmark~Yes & \xmark~No & Latent direction discovery \\
\hline

\textbf{Haas et al. \cite{haasDiscoveringInterpretableDirections2023}} (unsupervised) & H-space & Unsupervised PCA/Jacobian analysis & \cmark~Yes & \cmark~Yes & \xmark~No & Latent direction discovery \\
\hline

\textbf{Haas et al. \cite{haasDiscoveringInterpretableDirections2023}} (supervised); \newline \textbf{Li et al. \cite{liSelfDiscoveringInterpretableDiffusion2023}} & H-space & Supervised attribute classifiers & \cmark~Yes & \xmark~No & \cmark~Yes & Image editing / debiasing \\
\hline

\textbf{Parihar et al. \cite{pariharBalancingActDistributionGuided2024}} & H-space & Attribute Distribution Predictor (ADP) using pseudo-labeled attributes & \cmark~Yes & \xmark~No & \cmark~Yes  & Image editing / debiasing \\
\hline

\textbf{Shi et al. \cite{shiDissectingMitigatingDiffusion2025}} & H-space & Sparse autoencoder with gradient attribution & \cmark~Yes  & \xmark~No & \cmark~Yes & Image editing / debiasing \\

\hline

\textbf{SCALEX (Ours)} & H-space & Prompt-aligned concept vectors, extracted zero-shot & \cmark~Yes & \cmark~Yes & \cmark~Yes (prompt defines direction) & Large-scale bias analysis and knowledge discovery \\

\end{tabular}
\caption{\vspace{-10pt}Comparison of methods for discovering and applying semantic directions in diffusion models.\vspace{-9pt}}
\label{tab:method-comparison}
\end{table*}

\subsection{Bias and Fairness Evaluation}
\vspace{-2pt}
Text-to-image diffusion models often reinforce societal biases, particularly around gender and race. Many recent works evaluate these patterns using controlled prompts or attribute classifiers. Wu et al. \cite{wuRevealingGenderBias2025} use prompt triplets to quantify gender disparities, while the T2ISafety benchmark \cite{liT2ISafetyBenchmarkAssessing2025} provides a broader evaluation suite across fairness, toxicity, and privacy. However, these methods are limited to known demographic attributes and predefined taxonomies. They lack the capacity to uncover more subtle or unanticipated patterns in model behavior. Orgad et al. \cite{orgadEditingImplicitAssumptions2023} illustrate the presence of such implicit assumptions, such as default color or profession associations, which are difficult to detect using fixed taxonomies.

Several works aim to mitigate bias via prompt edits or model interventions. However, all rely on knowing which biases to correct, and are therefore limited to the same predefined categories and demographic axes used during evaluation. In contrast, our work addresses this limitation by focusing on scalable bias identification rather than predefined evaluation. This allows us to uncover structural patterns in the model's behavior without prior assumptions about which biases to test.

\subsection{Semantic Directions}
\vspace{-2pt}

Many existing methods rely on semantic directions: vectors in latent space that correspond to human-interpretable concepts. This idea, popularized in word embeddings \cite{mikolovLinguisticRegularitiesContinuous2013, mikolovEfficientEstimationWord2013}, has been widely adopted in generative models. In GANs, semantic directions enable controllable image editing \cite{radfordUnsupervised2015}, such as modifying the face \cite{shenInterpretingLatentSpace2020} or pose \cite{jahanianSteerabilityGenerativeAdversarial2019, plumeraultControllingGenerativeModels2020}, as well as analyze memorability \cite{goetschalckxGANalyzeVisualDefinitions2019} and interpolate between images \cite{voynovUnsupervisedDiscoveryInterpretable2020, zhuInDomainGANInversion2020, cherepkovNavigatingGANParameter2021}

\subsection{Semantic Directions in Diffusion Models}
\vspace{-2pt}
Extending semantic directions to diffusion models is more challenging due to their iterative structure and multiple conditioning inputs. Some approaches operate in text embedding space \cite{kimRethinkingTrainingDebiasing2025, friedrichFairDiffusionInstructing2023}, modifying prompt vectors to steer generation. While effective for editing, these directions primarily reflect language priors and offer limited visibility into the model’s internal latent structure.

Other methods apply classifier-driven guidance or manipulate latent subspaces during generation \cite{brackSEGAInstructingDiffusion2023, wangConceptAlgebraScoreBased2023}. These require per-timestep recomputation and do not yield reusable, interpretable vectors.

A separate line of work aims to discover latent directions directly. NoiseCLR \cite{dalvaNoiseCLRContrastiveLearning2023} learns attribute-aligned vectors through contrastive learning, while unsupervised methods use Jacobian analysis or PCA to extract dominant axes \cite{parkUnsupervisedDiscoverySemantic2023, haasDiscoveringInterpretableDirections2023}. Although these approaches reveal influential directions, they rely on manual inspection to assign semantic meaning, limiting their scalability for large-scale analysis.

\vspace{-2pt}
\subsection{H-space}
\vspace{-2pt}
The middle layer of the U-Net, known as H-space, provides a high-dimensional, linear, and stable representation well-suited for semantic analysis \cite{kwonDiffusionModelsAlready2022}. It has been used for unsupervised concept discovery \cite{parkUnsupervisedDiscoverySemantic2023}, attribute editing \cite{liSelfDiscoveringInterpretableDiffusion2023, haasDiscoveringInterpretableDirections2023}, fairness-aware generation \cite{pariharBalancingActDistributionGuided2024}, and neuron-level bias mitigation \cite{shiDissectingMitigatingDiffusion2025}. However, these methods often rely on training classifiers, optimizing attributes, or manually inspecting discovered features.

In contrast, \textbf{SCALEX} enables scalable, zero-shot exploration of H-space by aligning latent directions with textual prompts. Table \ref{tab:method-comparison} summarizes key differences across methods. SCALEX uniquely offers:

\begin{itemize}
    \item \textbf{Fully Automated \& Zero-Shot:} No attribute-specific training required.
    \item \textbf{No Manual Interpretation:} Prompt-aligned directions eliminate post hoc labeling.
    \item \textbf{Reusable Vectors:} Stable across seeds and timesteps.
    \item \textbf{Large-Scale Analysis:} Supports broad conceptual and bias exploration across hundreds of prompts.
\end{itemize}

\vspace{-3pt}
\section{Methodology}\label{sec:method}  
\vspace{-3pt}
Our goal is to enable scalable and automated analysis of bias and latent concept structure in diffusion models. This requires two crucial steps. First, a latent representation that is meaningfully tied to natural language concepts, eliminating manual interpretation. Second, it must allow prompt-aligned directions to be extracted directly without training or concept-specific supervision.

\subsection{Prompt-Aligned Latents in H-space}
\vspace{-2pt}
Concept Algebra \cite{wangConceptAlgebraScoreBased2023} constructs editing directions from prompt-conditioned denoised outputs, which are inherently tied to text and require no manual labeling. However, these vectors are tied to specific noise seeds and timesteps, making them non-reusable and unsuitable for systematic bias analysis.

To overcome this, we operate in \textbf{H-space}, the activation space of the middle U-Net layer. H-space has been shown to capture consistent, semantically meaningful features across prompts and noise conditions \cite{kwonDiffusionModelsAlready2022, parkUnsupervisedDiscoverySemantic2023}, and supports operations such as vector arithmetic and clustering.

\begin{figure}
    \centering
    \includegraphics[width=1\linewidth]{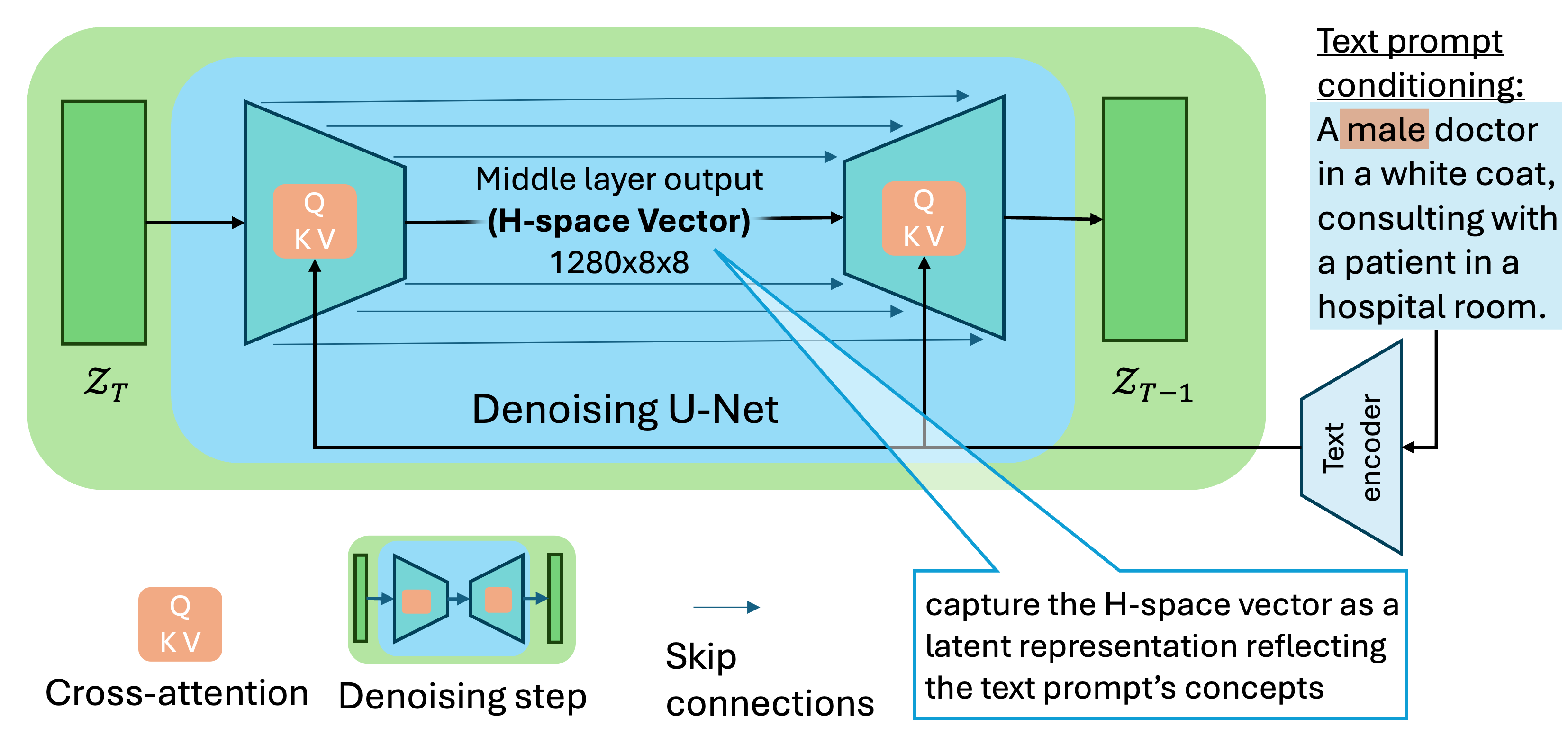}
    \caption{Vectors are captured from the H-space of the model, while conditioned on some text prompt. \vspace{-10pt}}
\label{fig:process}
\end{figure}

We extend this space by aligning each H-space vector to a natural language prompt. Given a prompt, we condition the diffusion model and extract the corresponding H-space activation during the forward pass (see Figure \ref{fig:process}). This yields a reusable, prompt-aligned latent vector. This direct linkage between text inputs and latent vectors enables systematic, large-scale analysis as each latent vector no longer requires manual interpretation to identify its meaning.

\subsection{Stabilizing the Latent Space with LCM}

Although H-space offers a semantically meaningful representation, activations observed directly during standard diffusion generation are noisy and highly dependent on the denoising timestep. Different timesteps encode different semantic scales \cite{parkUnsupervisedDiscoverySemantic2023}. Earlier timesteps influences coarser attributes while later timesteps focus on high-frequency components. Even though trained H-space directions can remain consistent across timesteps for targeted editing \cite{liSelfDiscoveringInterpretableDiffusion2023}, directly extracting prompt-aligned vectors without supervision results in unstable and entangled representations.

To obtain a stable representation that integrates semantic information across the full denoising trajectory, we apply \textbf{Latent Consistency Models (LCMs)} \cite{luoLatentConsistencyModels2023} through the latent consistency model LORA \cite{luoLCMLoRAUniversalStableDiffusion2023} (LCM-LORA).
Latent Consistency Models directly predict the solution of the Probability Flow ODE (PF-ODE) at \( t = 0 \), based on a consistency function \( f_\theta(z, c, t) \) that maps from a noisy sample at any timestep \( t \) to its denoised state.
This allows efficient extraction of prompt-aligned H-space activations in a single forward pass.

To improve image quality, LCM samplers add noise back into the image after the first prediction and denoise again to achieve a final image after 2-6 iterations. 
For our purposes, we discard the subsequent sampling steps used for image refinement and retain only the H-space vector from the initial prediction. This enables efficient, one-shot extraction of semantic directions aligned to text.


\begin{figure*}
    \centering
      \subfloat[]{
      \includegraphics[width=.8\textwidth]{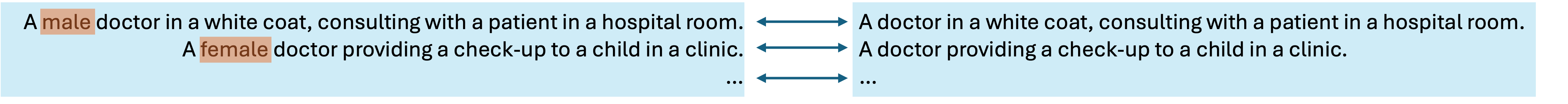}
          \label{fig:one-to-one}
      } \\
      \centering
      \subfloat[]{
      \includegraphics[width=.8\textwidth]{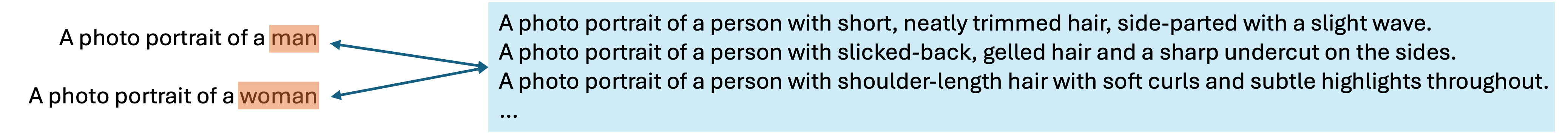}
          \label{fig:one-to-many}
      } \\
      \centering
      \subfloat[]{
      \includegraphics[trim={0 1.5cm 0 1cm},width=.8\textwidth]{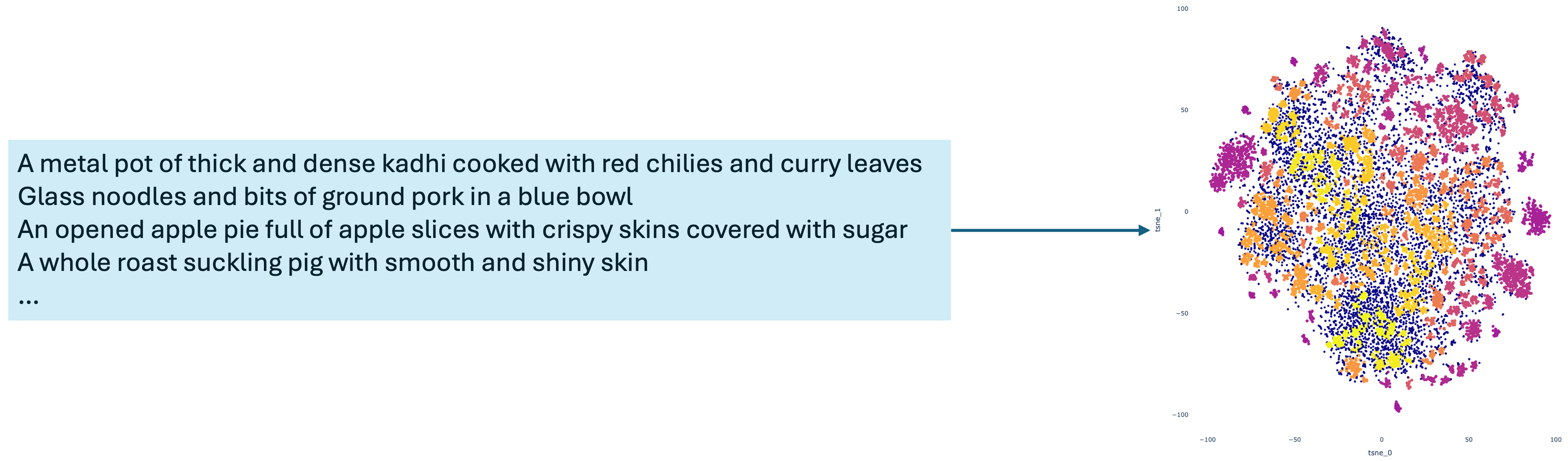}
      \label{fig:clustering}
      }
    \caption{Overview of the three experimental approaches for analyzing H-space representations. (a) Identify "defaults" through one-to-one comparisons, isolate the impact of adding or removing a single concept. (b) Find detailed trait descriptors for categories through one-to-many comparisons, which rank multiple prompts along a semantic spectrum. (c) Clustering captures broader patterns and associations across diverse captions beyond predefined categories. \vspace{-0.5cm}}
    \label{fig:three-experiments}
\end{figure*}

A detailed discussion of how semantic directions discovered with the LCM-LORA transfer to the standard latent diffusion model (LDM) is provided in the Supplementary Material (Section \ref{sec:lcm-to-ldm}).

\subsection{Automated Bias Analysis}

With a stable, text-aligned latent representation in hand, we design analyses to detect and quantify biases in diffusion models. We structure this as a pipeline that reflects both the types of datasets typically available and the different granularity of bias analysis. We first establish the model’s defaults (e.g., whether the unmarked prompt ‘a doctor’ aligns more with ‘male’ or ‘female’). We then refine these comparisons with descriptors, identifying which specific attributes the model associates with each option. Building on this, we explore clustering, which reveals emergent conceptual structures beyond predefined categories. Finally, we validate the semantic interpretability of these vectors via conditioning in image generation.


\subsubsection{Defaults: One-to-one comparison}

One-to-one comparisons occur between prompts that either include or exclude a specific concept. This measures how closely the ``default'' expression of the original prompt aligns with the prompt containing the additional concept. For example, the original prompt of ``a doctor'' may be compared against ``a female doctor'' or ``a male doctor''. By computing cosine distances between these paired vectors, we quantify the degree to which the presence of certain words affects the model's internal concept representation.

This setup parallels the triplet-prompt framework used by Wu et al. \cite{wuRevealingGenderBias2025}, which compares between images generated with gendered and neutral prompts and analyze bias across text embeddings, cross-attention maps, denoised latents, and image outputs. In contrast, we operate in H-space, which is more stable than denoised latents \cite{kwonDiffusionModelsAlready2022} and avoids reliance on external object detectors.

\subsubsection{Descriptors: One-to-many comparison}
\label{sec:one-to-many-methods}

In many cases, it is infeasible to modify every sentence to test the impact of adding a particular word. Instead, similar to the sentence embedding association test (SEAT) \cite{maySEATMeasuringSocialBiases2019}, we can compare a list of descriptor prompts to concepts of interests. Through measuring the cosine distance of each descriptor to the concepts of interest, we can rank the descriptors as closer to one concept or the other. 

In this setup, the cosine distance cannot be used directly to rank the descriptors. This is because we are only interested in the relative distance of each descriptor to the target concepts. For example, suppose one descriptor created a black-and-white image while the target concepts produced colored images. This descriptor will be further away from all target concepts compared to other descriptors, but may still be significantly closer to one target concept compared to another.

Since direct cosine distances are affected by global positioning in latent space, we normalize distances relative to the mean of all target concepts as follows:

\vspace{-6pt}
\begin{equation}
    d'_{i,j} = d_{i,j} - \frac{\sum_{k \in [n]}{d_{i, k}}}{n}
\end{equation}
\vspace{-8pt}

\noindent where $d_{i,j}$ is the cosine distance between the $i^{th}$ descriptor and the $j^{th}$ concept, and $n$ is the number of target concepts. In the case where there are only 2 target concepts, it is also sufficient to subtract the distance of one concept from the other.


Other normalization methods, such as projecting into a subspace defined by the target concepts or additionally dividing by standard deviation, are discussed in the supplementary materials.

After ranking the descriptors, it is also possible to use a LLM to summarize the characteristics that are more likely to occur for each target concept. This automation allows for evaluating biases at a much larger scale than prior methods.

\subsubsection{t-SNE visualization and clustering}

The above two use cases both involve targeted analysis of bias for a particular concept or set of concepts. Yet, it is often the case that unknown biases exist. Through obtaining the H-space vectors from a dataset of naturally occurring captions, we can use t-SNE \cite{maatenTSNE2008} to visualize the high-dimension data in 2D space. Clustering algorithms such as HDBSCAN \cite{campelloDensityBasedClusteringBased2013,campelloHierarchicalDensityEstimates2015,mcinnesAcceleratedHierarchicalDensity2017} can also be used to automatically group points to form clusters. This allows for an unsupervised exploration of how the model organizes concepts internally. 

Each cluster can also be summarized using a LLM. This provides a clear visualization of which concepts the diffusion model considers more important than others. Outliers within a cluster can also give a clue on how the model differentiates similar concepts, highlighting edge cases where the model's learned representations deviate from expected patterns or mix multiple concepts together.

Distances between clusters can also be used to provide insight on which concepts the model considers to be similar to each other.

When captions are labelled with categories, we can also calculate cluster overlap to evaluate if the diffusion model is likely to confuse one class with another. 

Overall, this allows for broad and automated exploration of conceptual organization.



\subsection{Image Conditioning}

To further validate the H-space vectors that were obtained, we explore their use in direct image manipulation. By applying H-space vectors as conditioning factors, we can modify the generated images in a controlled manner. Specifically, we take an averaged H-space vector from a particular concept (e.g., a cluster representing "square plates" or "curly hair") and add it to the H-space of new image generations. This enables the diffusion model to generate images that align with the semantic attributes associated with that vector. The effectiveness of this approach is demonstrated through conditioning experiments (Figure \ref{fig:hspace_image_edit} and \ref{fig:conditioning_hspace_food}).


\section{Experiments/Analysis}\label{sec:experiments}

This section presents the experimental results obtained using the proposed framework for scalable bias discovery in diffusion models. We evaluate biases in latent representations through finding defaults, refining with descriptors, and cluster-based analysis beyond pre-defined categories. All results in this section use the Stable Diffusion 1.5 model unless otherwise noted. Experiments are replicated using the Stable Diffusion XL (SDXL) model in the supplementary (Section \ref{sec:sdxl}).

\subsection{Defaults: One To One Comparison}\label{sec:one-to-one-experiment}


\begin{figure}
    \centering
    \includegraphics[width=0.8\linewidth]{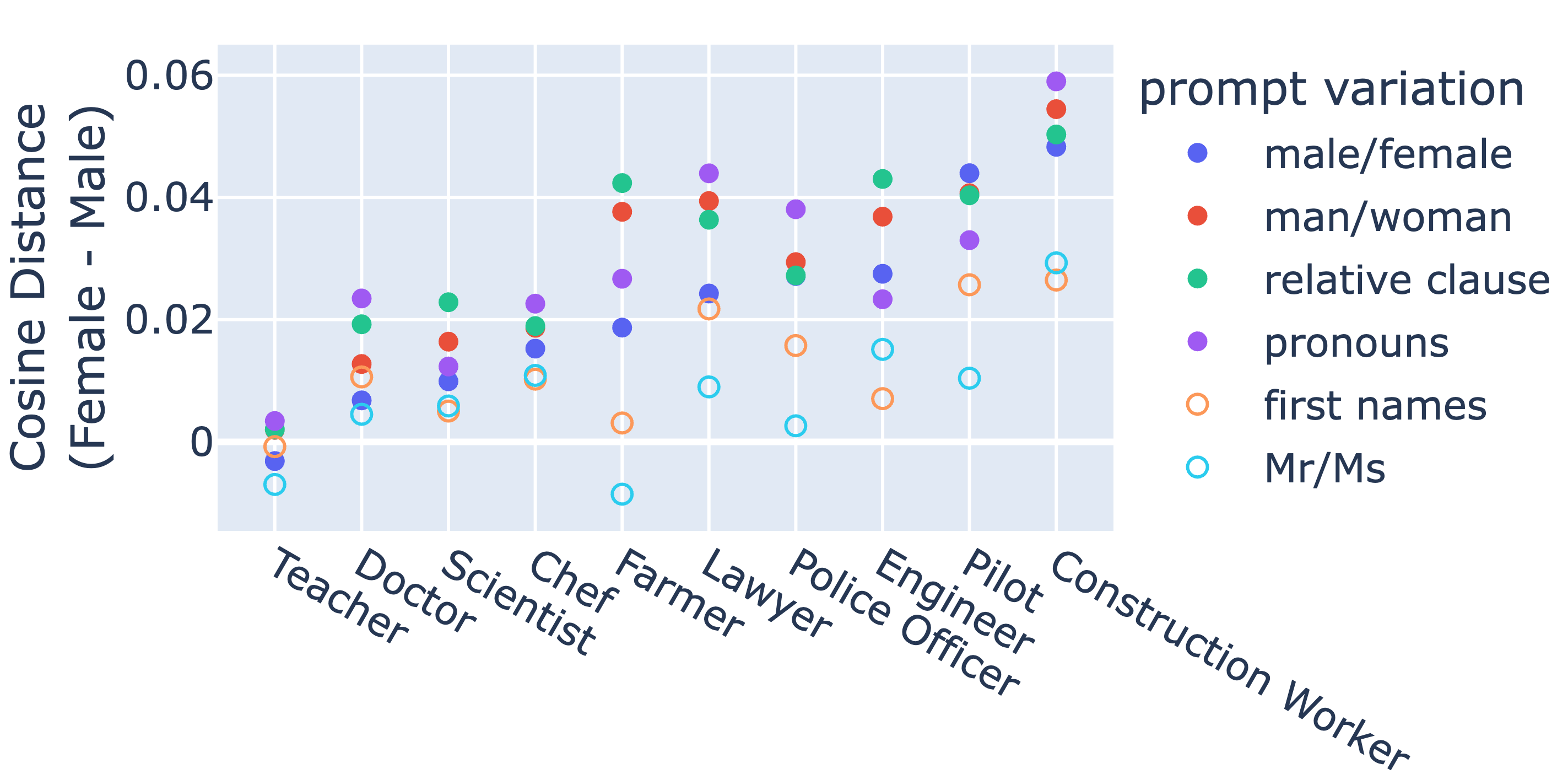}
    \caption{For each profession, we plot the average difference in male and female cosine distances for five prompt variants:
  \emph{female/male} (baseline), \emph{woman/man}, relative clause
  (\emph{a doctor who is a woman/man}), pronouns
  (\emph{she/he is a doctor}), first names (\emph{Sarah/John, a doctor}),
  and honorifics (\emph{Ms./Mr.\ surname}). Results are strongly correlated with the baseline \emph{female/male} phrasing (Pearson $r = 0.92, 0.86, 0.84$ respectively for \emph{man/woman}, \emph{relative clause}, \emph{pronouns}). First names ($r=0.85$) and honorifics ($r=0.72$) show positive but weaker alignment due to (i) additional priors like age or region embedded in names, and (ii) honorifics being weaker gender signals in caption distributions.  See Supplementary for more detailed analysis of their differences.\vspace{-8pt}}
    \label{fig:gender_prompt_variation}
\end{figure}

In this experiment, we quantify gender bias across professions. For each profession, we first generate neutral descriptors (see Section \ref{sec:professions_prompts} in supplementary). We then add gender markers in both directions (e.g. a male doctor, a female doctor). H-space vectors are extracted for all prompts. 


To test robustness, we also analyze five alternative phrasings: man/woman, relative clause (a doctor who is a woman/man), pronouns (she/he is a doctor), first names (Sarah/John, a doctor), and honorifics (Mr./Ms. surname). 


We pair each neutral prompt with its male and female variants and compute the difference in cosine distance to the neutral vector:  
\begin{equation}
    \Delta = d(h_f, h_0) - d(h_m, h_0),
\end{equation}
\noindent where $h_f$ and $h_m$ denote the H-space vectors for the female and male prompts, and $h_0$ the neutral prompt. Positive values indicate that the neutral prompt aligns more closely with the male representation.  

The distance between gendered and non-gendered prompts can reflect more than gender alone. Extra traits (e.g., style, age, or background) may appear. When they occur sporadically, averaging across seeds reduces their effect. When they occur consistently, we treat them as learned associations rather than noise. Thus SCALEX measures whether the default concept (e.g., doctor) aligns more with one gendered option and its correlated attributes. Descriptors analysis in Section \ref{sec:one-to-many-experiment} then isolates which traits drive these associations.

Figure~\ref{fig:gender_prompt_variation} visualizes this metric across all professions and prompt variations. Professions such as \emph{chef}, \emph{construction worker}, and \emph{police officer} consistently default closer to male, while \emph{teacher} appears more balanced.




We validate these biases by classifying the images generated from non-gendered prompts using CLIP \cite{radfordCLIP2021} as a zero-shot classifier. Professions favoring the male representation in H-space correspond to the lowest probability of generating female-presenting images (Table~\ref{tab:professions_percent_female}), confirming that divergence in H-space reflects generative bias.

\begin{table}[ht]
\footnotesize
\centering
\begin{tabular}{|p{2.6cm}|p{1cm}|p{1cm}|p{0.95cm}|p{0.95cm}|}
\hline
\multirow{2}{*}{\textbf{Profession}} & \multicolumn{2}{p{2.1cm}|}{\textbf{\% female images}} & \multicolumn{2}{p{1.9cm}|}{\textbf{Cosine Distance (Female - Male)}} \\ \cline{2-5}
 & \textbf{SD1.5} & \textbf{SDXL} & \textbf{SD1.5} & \textbf{SDXL} \\ \hline
\textbf{Teacher}             & 58.33 & 60.00 & -0.0003 & 0.0107 \\
\textbf{Scientist}           & 41.67 & 33.33 & 0.0088  & 0.0172 \\
\textbf{Chef}                & 11.67 &  3.33 & 0.0138  & 0.0185 \\
\textbf{Lawyer}              & 25.00 & 10.00 & 0.0145  & 0.0398 \\
\textbf{Farmer}              & 23.33 & 10.00 & 0.0153  & 0.0472 \\
\textbf{Doctor}              & 45.00 & 23.33 & 0.0170  & 0.0371 \\
\textbf{Police Officer}      & 10.00 &  6.67 & 0.0295  & 0.0616 \\
\textbf{Engineer}            &  1.67 &  6.67 & 0.0394  & 0.0428 \\
\textbf{Construction Worker} &  1.67 &  0.00 & 0.0428  & 0.0736 \\
\textbf{Pilot}               &  1.67 &  0.00 & 0.0470  & 0.0917 \\ \hline
\end{tabular}
\caption{Comparison of gender bias across professions for Stable Diffusion 1.5 (SD1.5) and Stable Diffusion XL (SDXL). For both models, larger cosine distances of female H-space vectors compared to male correspond to lower percentages of female-presenting images generated from non-gendered prompts. The trend is consistent across architectures, though SDXL exhibits slightly higher bias magnitudes for several professions. 
}
\label{tab:professions_percent_female}
\end{table}

\subsection{Descriptors: One to many comparison}
\label{sec:one-to-many-experiment}

In the following example, ChatGPT 4o \cite{openai2024chatgpt} was used to generate gender-neutral descriptions of facial portraits. H-space vectors were then collected using these captions. Those results were then compared with H-space vectors generated for each gender, namely: ``a photo portrait of a man'' and ``a photo portrait of a woman''. The difference in cosine distance between each gender-neutral prompt and the two gendered prompts was then used to rank the gender-neutral prompts from ``most similar to man'' to ``most similar to woman'' (Table \ref{tab:hair_male_female}).

\begin{table}
\footnotesize
    \centering
    \begin{tabular}{|p{4.8cm}|p{2.2cm}|}
    \hline
         caption& Cosine Difference (Female - Male) \\ \hline 
         A photo portrait of a person with a bald head and a light sheen, complemented by a well-groomed beard.& 0.20140\\
         A photo portrait of a person with slicked-back, gelled hair and a sharp undercut on the sides.& 	0.11873\\
         ...& ...\\
         A photo portrait of a person with shoulder-length hair with soft curls and subtle highlights throughout.& -0.20819\\ \hline
    \end{tabular}
    \caption{A number of captions describing hair, ranked from closest to male to closest to female\vspace{-10pt}}
    \label{tab:hair_male_female}
\end{table}

Ranked cosine distances (Table 2) reveal that descriptors like 'bald with a well-groomed beard' align more with male vectors, while longer, textured hairstyles align with female vectors. LLM-based summaries confirm these latent gender associations. ChatGPT-4 provided the following summary:
\vspace{-10pt}
\begin{quote}
\small
Person A is more likely to have hairstyles associated with a cleaner, sharper, or minimalist look. Specifically, bald or buzz cut, slicked-back or neatly styled hair, and shorter, simpler haircuts. Person B is more likely to have more complex or varied hairstyles, particularly longer and styled in different ways. Characteristics associated with person B include longer hair, more intricate styling, textured and curly hair, and more highlights or coloring.
\end{quote}
\vspace{-5pt}

\begin{figure}
    \centering
    \includegraphics[trim={0, 2cm, 0, 0},width=\linewidth]{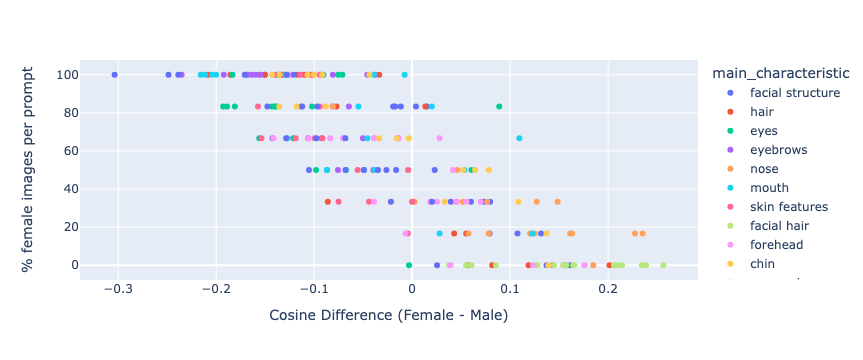}
    \caption{Correlation between the percentage of images classified as female (using CLIP) and the difference in cosine distances between gendered H-space vectors for each prompt. Prompts with higher cosine differences favouring female vectors are more likely to generate female-presenting images, indicating a strong association between H-space distances and perceived gender.}
    \label{fig:face_portrait_percent_female}
\end{figure}

Similar to the one-to-one comparisons in Section \ref{sec:one-to-one-experiment}, we also validate these findings using CLIP \cite{radfordCLIP2021} for gender classification. Figure \ref{fig:face_portrait_percent_female} plots the percentage of images classified as female against the difference in cosine distances of the H-space vectors, showing a strong correlation between the two metrics.

A similar comparison can also be made with multiple categories. For example, Table \ref{tab:hair_african} ranks hairstyles more likely associated with an ``African'' person than other races. The difference between ``African'' cosine distances and the mean cosine distance of all other races is used in this case. Other methods of normalizing the distance vector are included in the supplementary material.

\begin{table}
\footnotesize
    \centering
    \begin{tabular}{|p{4.9cm}|p{2.6cm}|}
    \hline
         caption& Cosine Difference (African - Mean(race)) \\ \hline 
         A photo portrait of a person with a curly afro, with tightly coiled hair forming a rounded shape.          & -0.23101\\
         A photo portrait of a person with a buzz cut and a clean-shaven look, highlighting a well-defined jawline. & 	-0.18972\\
         A photo portrait of a person with long, thick dreadlocks tied up into a half-up, half-down style.          & -0.18686\\
         ...& ...\\
         A photo portrait of a person with long, wavy hair with ombre coloring that fades from dark brown to light blonde.  &0.26944\\ \hline
    \end{tabular}
    \caption{A number of captions describing hair, ranked from closest to an ``African'' person relative to the mean distance to other races.\vspace{-10pt}}
    \label{tab:hair_african}
\end{table}

\subsection{Cluster Visualization}
\label{sec:cluster_visualization_experiment}

The following section uses the Food500-CAP \cite{maFood500CapFineGrained2023} dataset, which contains natural language captions for 24700 food images. This dataset was chosen as an example as food is a domain with high diversity in ingredients, composition, and regional biases, but is universally understood and rarely require any specialized knowledge to understand.

\begin{figure}
    \centering
    \includegraphics[trim={2cm 2cm 0.5cm 0.1cm},clip,width=\linewidth]{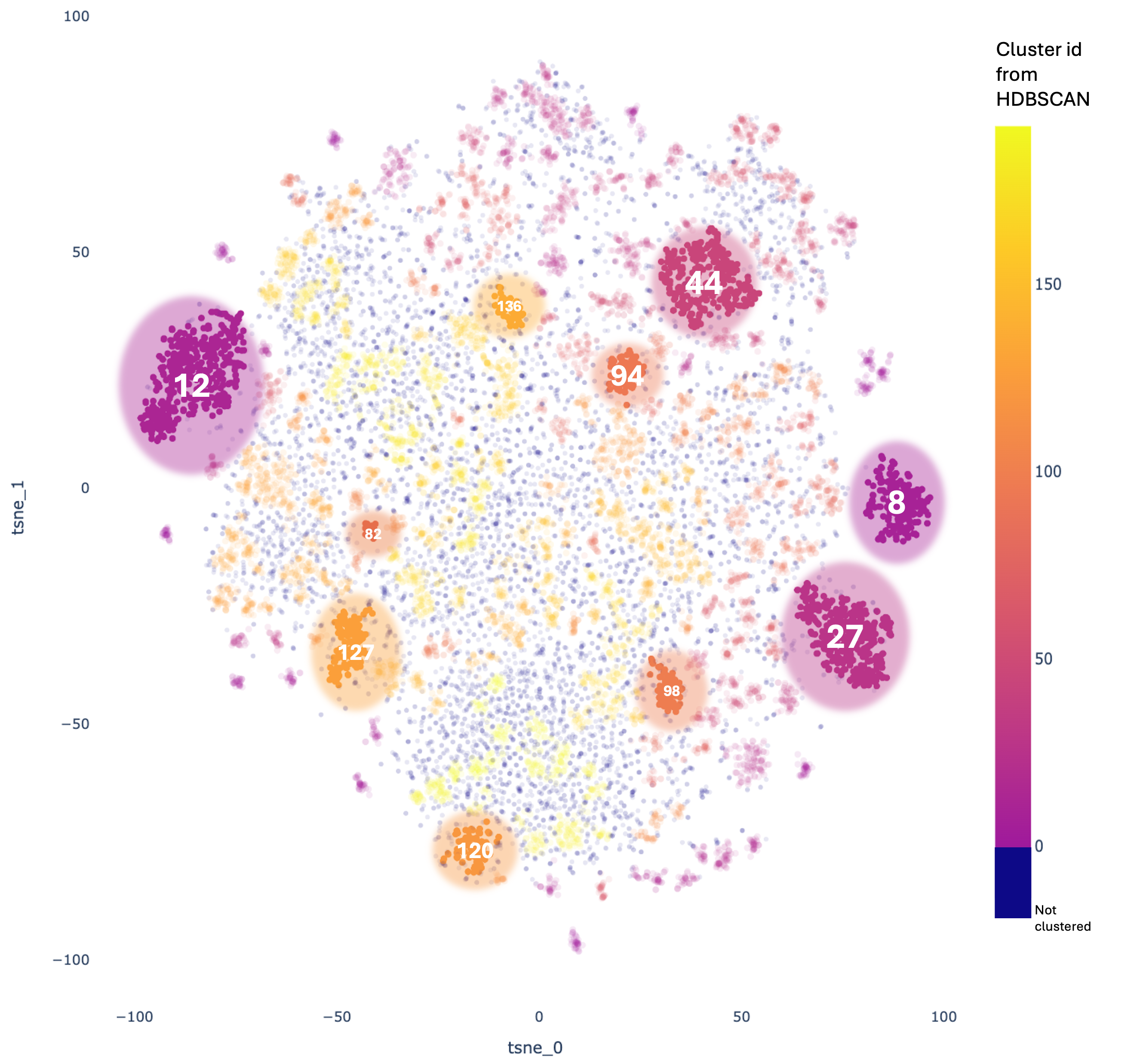}
    \caption{t-SNE visualization of H-space vectors obtained from the Food500-CAP dataset. Clusters are labelled based on the id assigned by HDBSCAN \cite{mcinnesAcceleratedHierarchicalDensity2017}. Some prominent clusters include (8) square dishes; (12) pots with soup; (27) sandwiches/bread; (44) pie; (82) seafood boil; (94) takeout boxes; (98) rectangular plates; (120) salads; (127) stir fried noodles; (136) purple vegetables, esp. purple cabbage\vspace{-0.5cm}}
    \label{fig:food-tsne}
\end{figure}

Figure \ref{fig:food-tsne} demonstrates a t-SNE \cite{maatenTSNE2008} visualization of H-space vectors obtained from this dataset and points out several prominent clusters.
Because each cluster is made up of captions instead of images, we can easily identify the common element in a cluster through LLMs such as ChatGPT \cite{openai2024chatgpt}. This removes the need to manually ``describe'' the latent vectors through observation, like Kwon \etal \cite{kwonDiffusionModelsAlready2022} and NoiseCLR \cite{dalvaNoiseCLRContrastiveLearning2023}.

Overall, the clusters are relatively within expectation, with soups being closely clustered with various noodles close by, and other sandwiches and burgers in their own clusters. However, there were also some surprises. For example, the only similarity between items in cluster 8 was the fact that the plate for the food was square. Yet another cluster (98) contained rectangular plates. This suggests that not only does the model place a lot of importance on geometry compared to perhaps the ingredients, but the model also makes a clear differentiation between a square and a rectangle.

\begin{figure}
    \includegraphics[trim={2cm 2cm 0.5cm 0.1cm},clip,width=.8\linewidth]{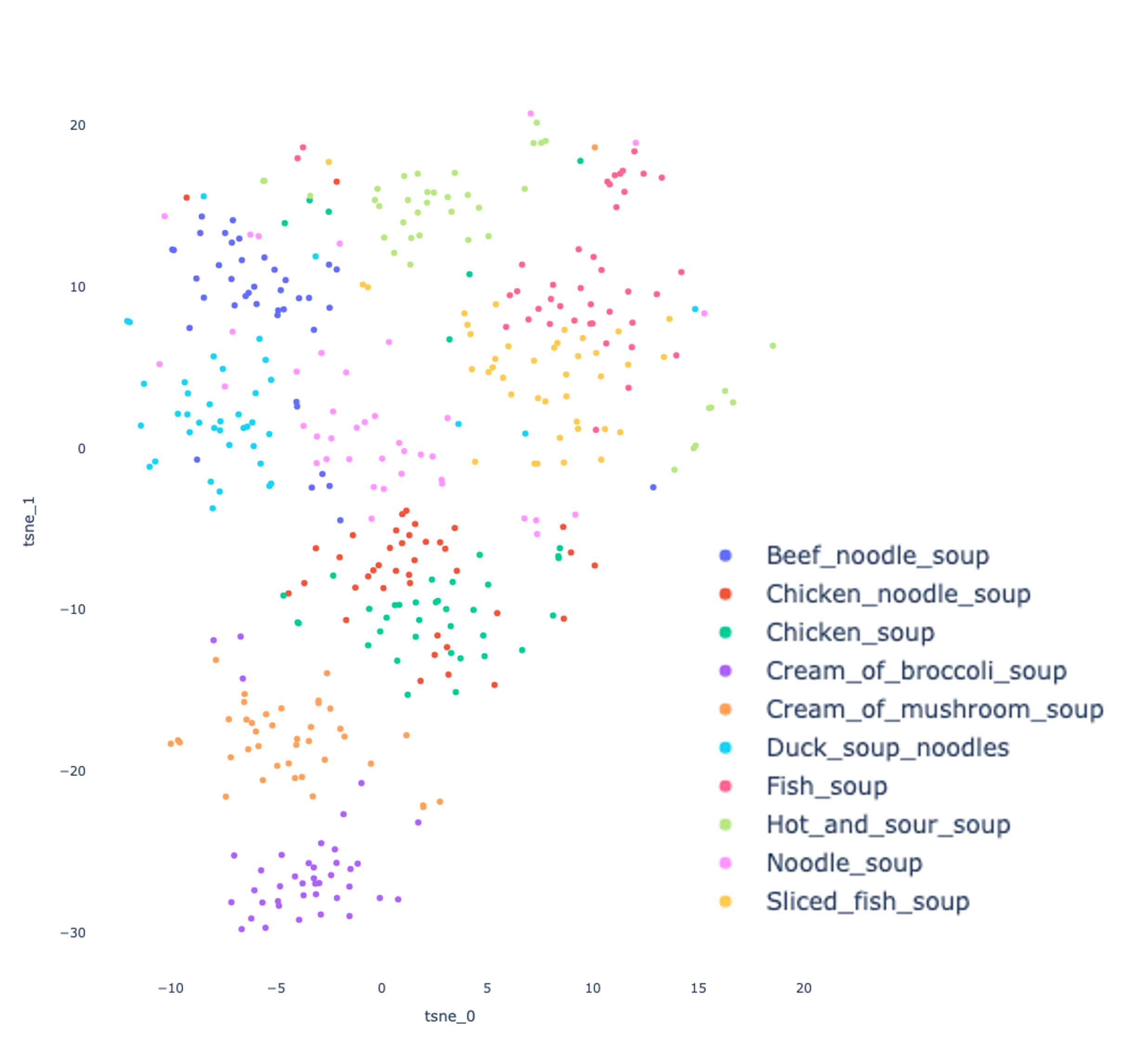}
    \caption{t-SNE visualization of H-space vectors obtained from the Food500-CAP dataset, limited to select soup categories. Strong overlaps, such as between “chicken soup” and “chicken noodle soup,” indicate where the model conflates related categories, while distinct separation (e.g., “cream of mushroom” vs. “cream of broccoli”) reveals sharper boundaries. Geographic clustering (Asian vs. Western soups) further suggests latent cultural associations. Overlap analysis thus provides a lens into how diffusion models internally encode and bias category relationships.\vspace{-0.5cm}}
    \label{fig:soup-tsne}
\end{figure}

To better understand how the diffusion model organizes food concepts, Figure \ref{fig:soup-tsne} isolates soup-related clusters, reducing visual clutter. The visualization highlights clear distinctions between certain categories, while others exhibit significant overlap, such as "chicken soup" and "chicken noodle soup." Such overlaps are not just visual curiosities but indicate latent biases in how categories are organized, reflecting ambiguity or conflation of culturally or socially meaningful concepts.”
Additionally, an emergent pattern aligns soups geographically, with Asian soups clustering towards the top and Western soups towards the bottom, hinting at a latent cultural representation in the model.

\begin{figure}
    \centering
    \includegraphics[width=.9\linewidth]{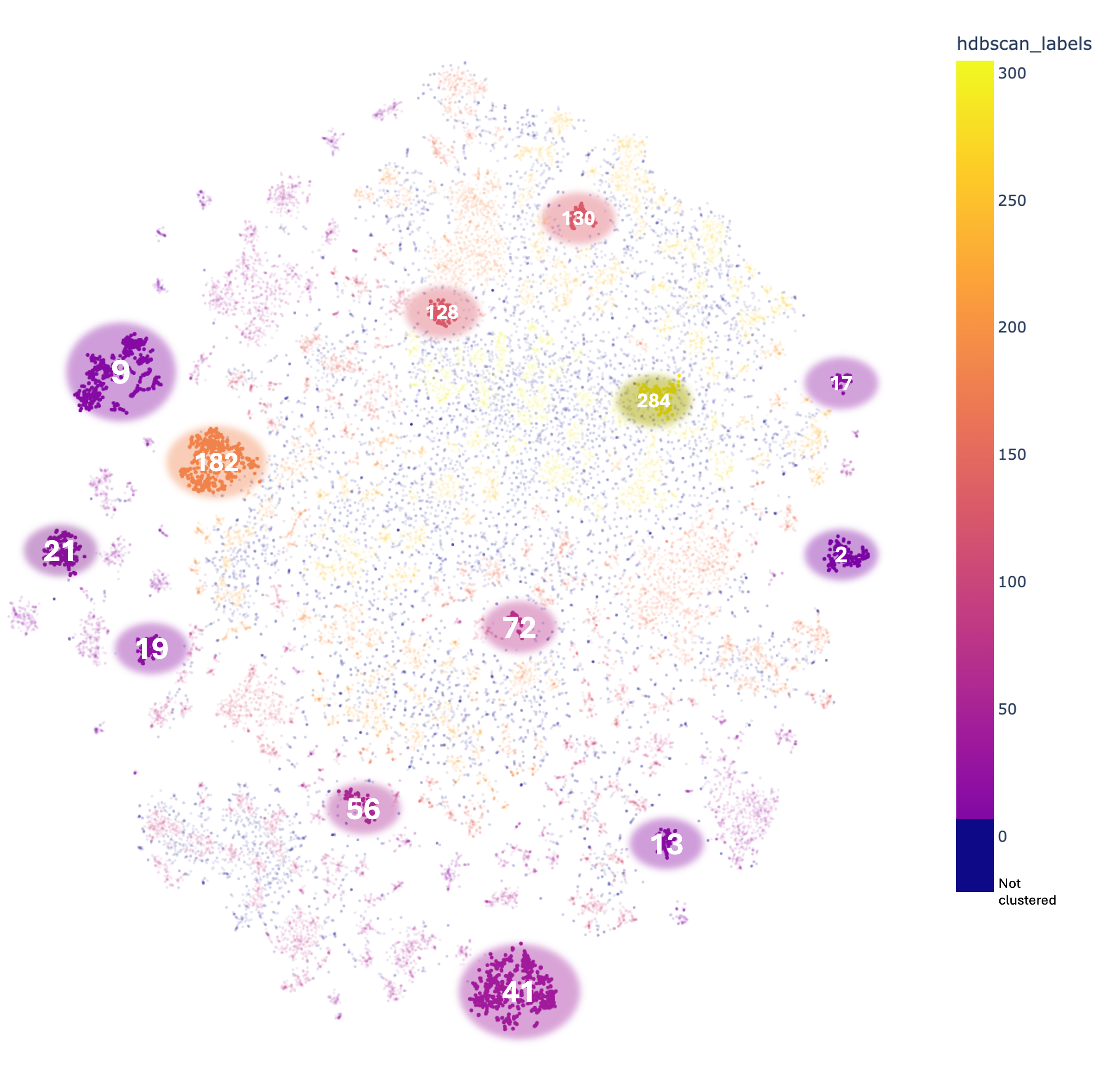}
    \caption{\vspace{-1pt}t-SNE visualization of H-space vectors obtained from a subset of the LAION dataset \cite{schuhmannLAION5BOpenLargescale2022}. Clusters are labelled based on the id assigned by HDBSCAN \cite{mcinnesAcceleratedHierarchicalDensity2017}. Some prominent clusters include (2) wedding; (9) phone case; (13) bathroom; (17) basketball; (19) glasses; (21) ring; (41) boots; (56) patterned fabric; (72) wallsticker; (128) poster; (130) magazine; (182) food; (284) group photo}
    \label{fig:laion-tsne}
\end{figure}

To further demonstrate the flexibility of this method, a randomly sampled subset of 500k captions from the LAION dataset \cite{schuhmannLAION5BOpenLargescale2022} is also visualized in Figure \ref{fig:laion-tsne}. 


This experiment demonstrates that even without explicitly targeting a specific concept, naturally occurring captions can reveal latent biases and associations. By leveraging clustering and visualization techniques, we can map out the conceptual landscape of the H-space, offering a broader understanding of how diffusion models encode and express complex ideas.

\subsection{Image Conditioning}

\begin{figure}
    \centering
    \includegraphics[width=.75\linewidth]{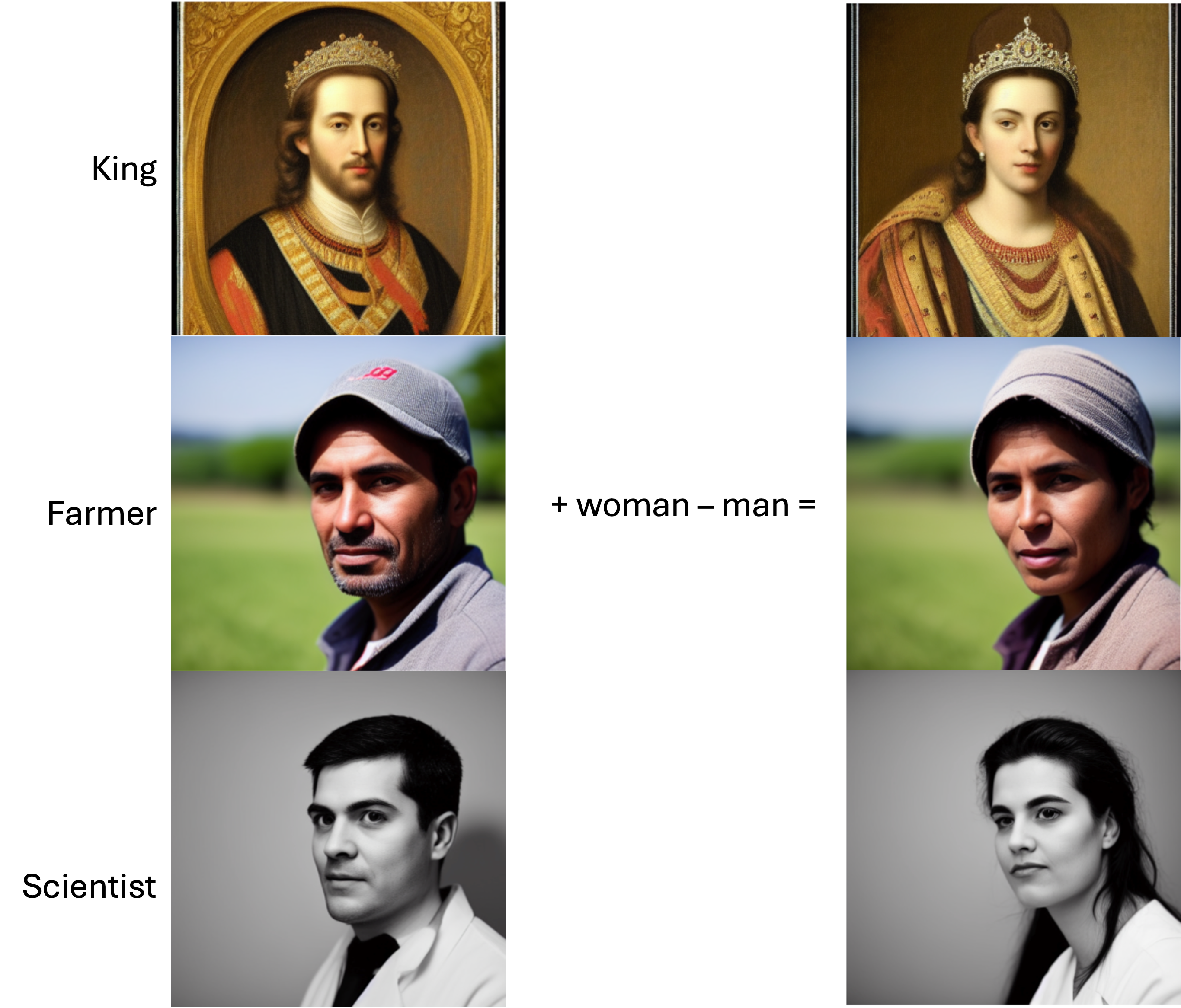}
    \caption{The transformation of a man to a woman is performed for both the classic example of Queen = King + woman - man as well as a Farmer and Scientist. H-space vectors are obtained for ``a portrait of a man'' and ``a portrait of a woman'' at the first timestep. That vector is then added at the H-space to achieve the image editing result.\vspace{-0.5cm}}
    \label{fig:hspace_image_edit}
\end{figure}

\begin{figure}
    \centering
    \subfloat[]{
        \includegraphics[trim={0 1.5cm 0 0},width=.6\linewidth]{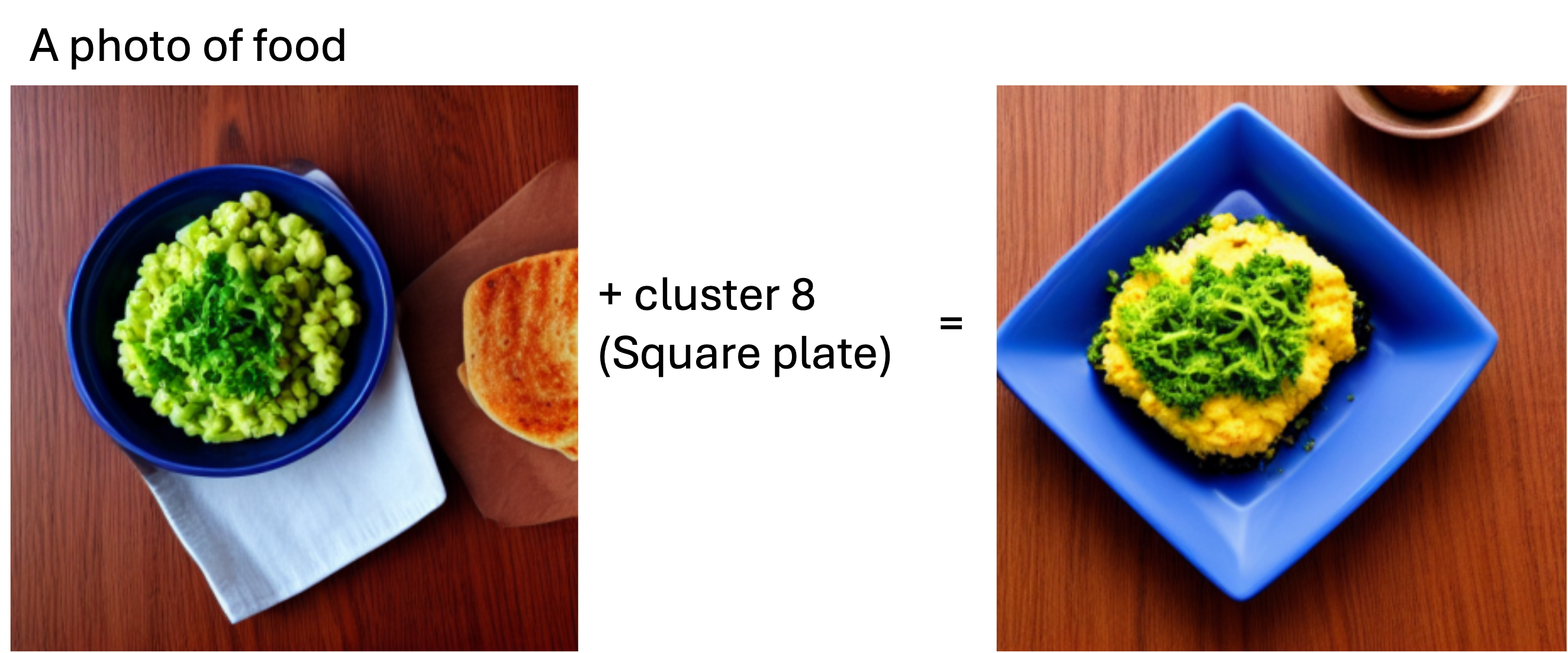}
        \label{fig:plate_0}
    }\\
    \subfloat[]{
        \includegraphics[trim={0 1.5cm 0 0},width=.6\linewidth]{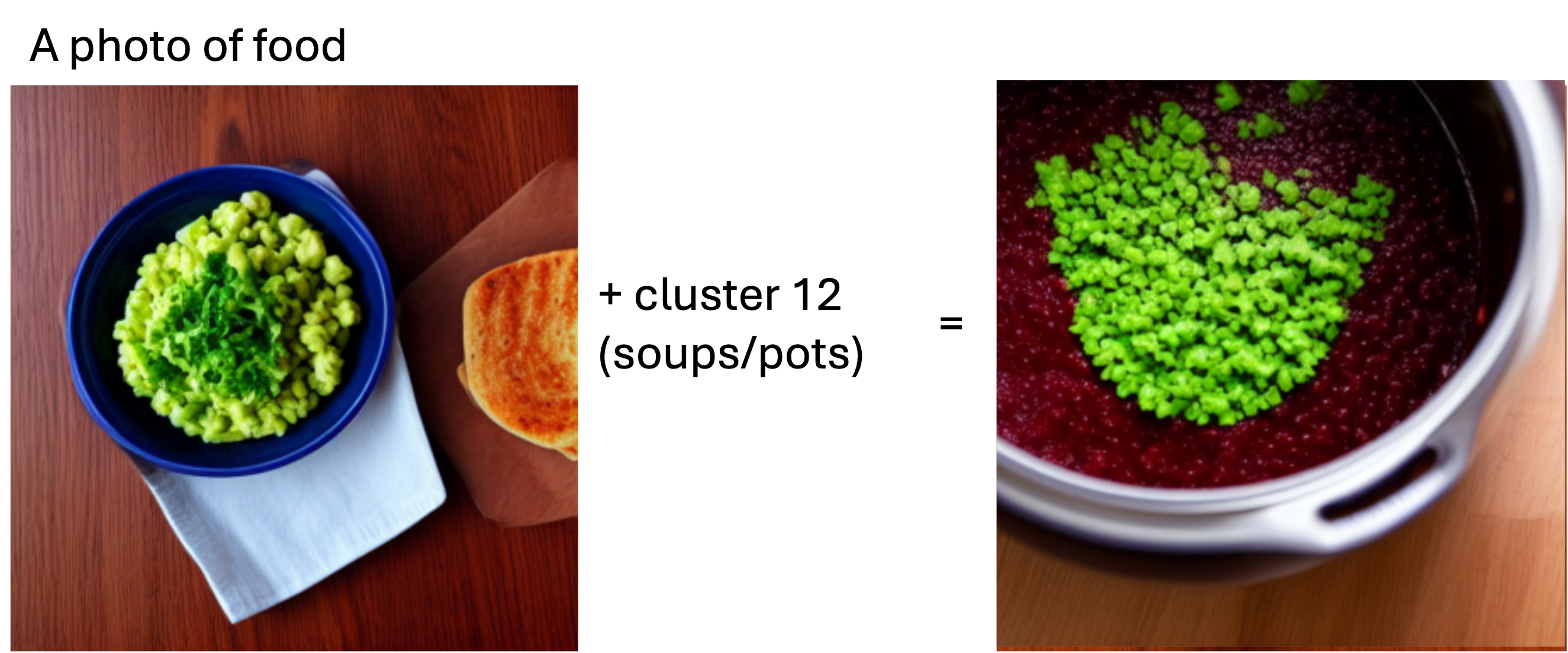}
        \label{fig:pot_0}
    }\\
    \subfloat[]{
        \includegraphics[trim={0 1.5cm 0 0},width=.6\linewidth]{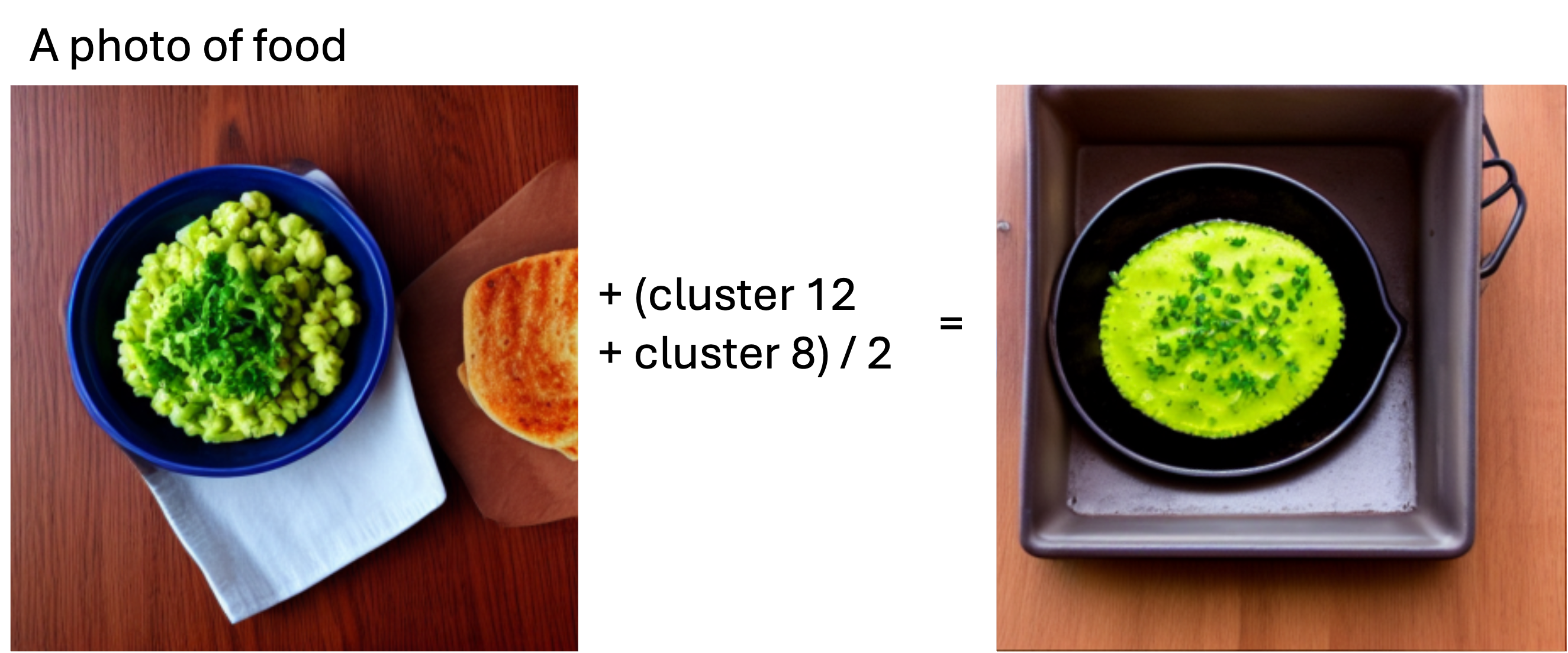}
        \label{fig:square_pot_0}
    }\\
    \subfloat[]{
        \includegraphics[trim={0 1.5cm 0 0},width=.6\linewidth]{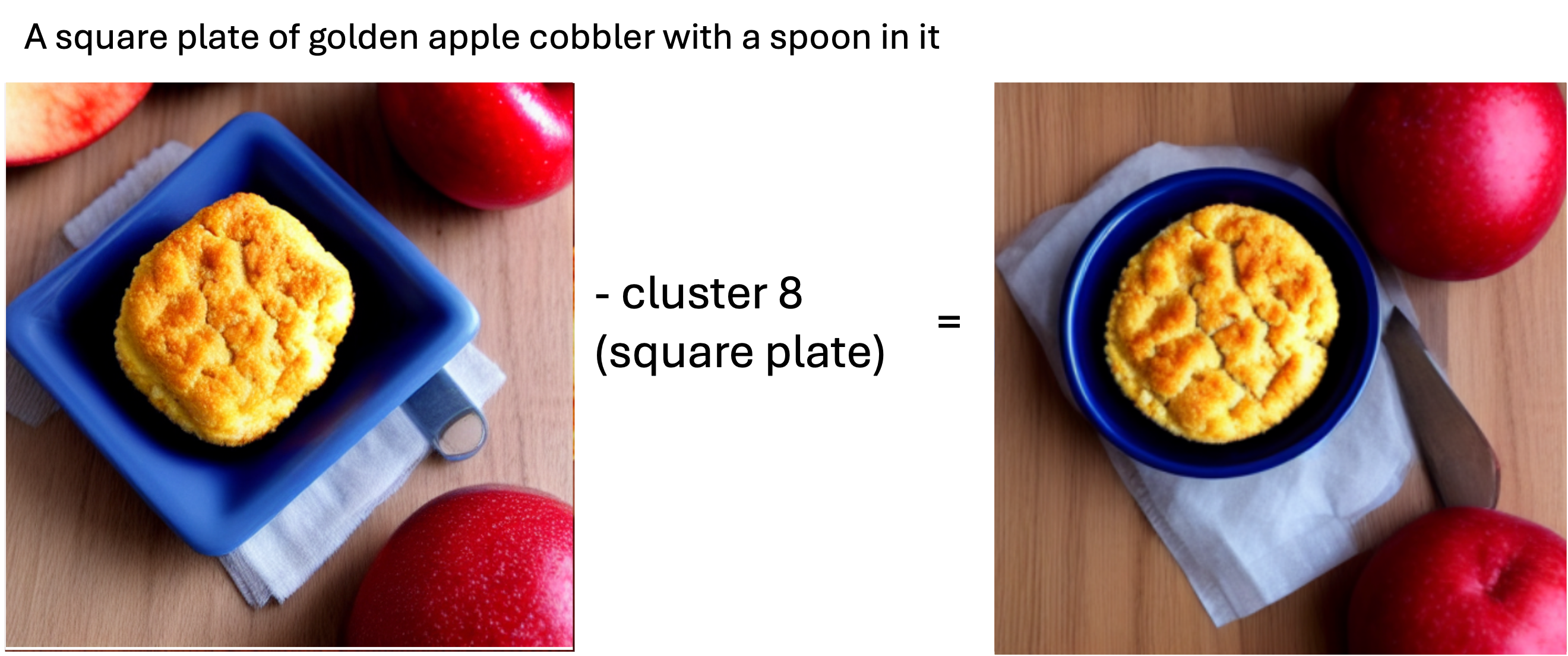}
        \label{fig:minus_square_plate_0}
    }
    \caption{Generated images conditioned on different H-space vectors. (a) and (b) show results using the average H-space vectors of individual clusters, such as "square plate" and "pot" categories. (c) demonstrates the effect of combining multiple cluster vectors, such as blending "square plate" and "pot" attributes. (d) visualizes the result of subtracting a cluster vector, removing characteristics associated with "square plate" while retaining other features. These results illustrate how combining and modifying H-space vectors enables controlled generation of conceptually meaningful outputs.\vspace{-1cm}}

    \hfill
    \label{fig:conditioning_hspace_food}
\end{figure}

To verify the H-space vectors in prior sections, we also use these vectors to condition the generated image. 

In Figure \ref{fig:hspace_image_edit}, a King, a farmer, and a scientist were all additionally conditioned by adding a H-space vector obtained from ``a portrait of a woman'' and subtracting a H-space vector obtained from ``a portrait of a man'' to achieve a gender-transformed version of the original prompt.

It is also possible to condition the image on multiple vectors, such as through taking the average of clusters from Section \ref{sec:cluster_visualization_experiment}. Example results of this can be seen in Figure \ref{fig:conditioning_hspace_food}. Different clusters can also be combined together, such as in Figure \ref{fig:square_pot_0}.

\section{Conclusion}

This paper introduced \textbf{SCALEX}, a framework for scalable, automated analysis of latent representations in diffusion models. By aligning H-space directions with natural language prompts, SCALEX enables zero-shot interpretation of arbitrary concepts, eliminating the need for classifier training or manual labeling. This approach allows large-scale comparison of model behavior across diverse semantic axes, far beyond what was possible with prior methods.

Our experiments show that SCALEX uncovers meaningful patterns in how diffusion models encode social biases, conceptual associations, and semantic groupings. These results highlight both the internal structure of the model's learned representations and the potential for automated tools to audit them at scale.

By making latent directions interpretable by design, SCALEX provides a practical foundation for understanding, evaluating, and eventually steering the behavior of generative models. While our method relies on the bottleneck structure of U-Net-based diffusion models, the analysis framework itself, including prompt-aligned comparisons, descriptor ranking, and unsupervised concept clustering, is broadly applicable. However, we acknowledge this architectural limitation. Future work will explore adapting SCALEX to transformer-based diffusion models and investigating whether analogous latent structures can support prompt-aligned concept analysis in these emerging architectures.
{
    \small
    \bibliographystyle{ieeenat_fullname}
    \bibliography{references}
}
\clearpage
\twocolumn[
\centering
\section*{Supplementary Material}
\addcontentsline{toc}{section}{Supplementary Material}
\vspace{1em} 
]

\appendix


\section{LCM-to-LDM H-space Alignment}
\label{sec:lcm-to-ldm}

Although the latent consistency model (LCM) refines the denoising process, its mid-layer representations remain conceptually similar to those of the standard latent diffusion model (LDM), particularly regarding semantic offsets used to manipulate concepts across timesteps. This similarity arises due to two main factors:

\begin{enumerate}
    \item Kwon et al. \cite{kwonDiffusionModelsAlready2022} demonstrate that small additive vectors applied in the U-Net's middle layer induce consistent semantic changes across different timesteps. These change vectors effectively encode semantic directions even under significant noise levels at early diffusion steps.
    
    \item During LCM training, the LoRA module is optimized to ensure the model produces the same final image—the probability flow ODE (PF-ODE) solution—across various sampling times. This training objective aligns the LCM's intermediate representations closely with those of the original LDM. Consequently, the same semantic directions (e.g., ``man → woman'') remain meaningful in both models because the LCM is explicitly trained to replicate the teacher LDM's outputs.
\end{enumerate}

\begin{figure}
    \centering
    \includegraphics[width=\linewidth]{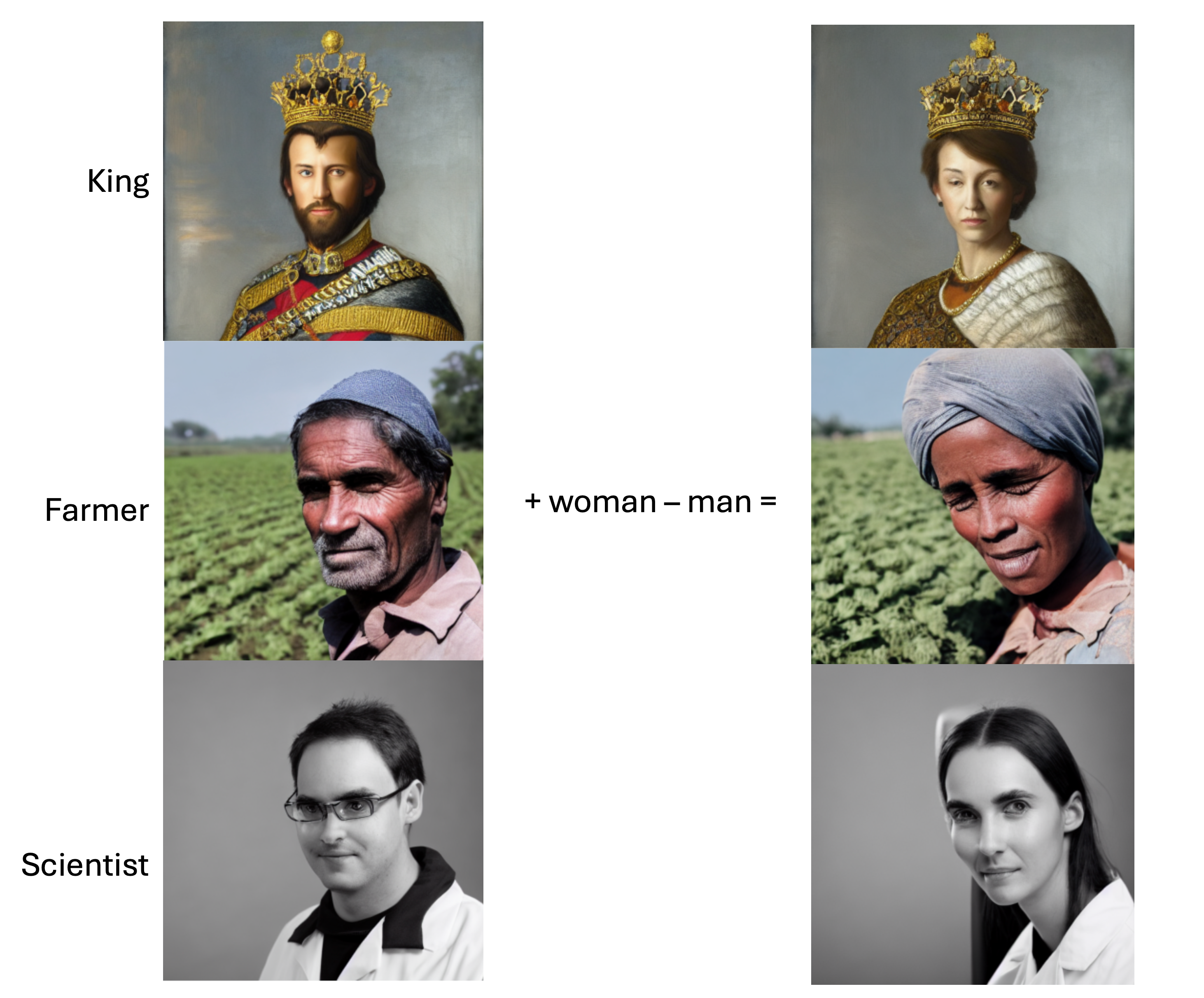}
    \caption{Applying semantic offsets derived from an LCM LoRA to an unmodified latent diffusion model (LDM). The transformations (e.g., ``male → female'') produce consistent changes in the generated images, confirming that semantic directions discovered in the LCM's H-space transfer effectively to the LDM.}
    \label{fig:ldm-transfer}
\end{figure}

To provide empirical support for this theoretical connection, we replicated our semantic offset experiments by applying vectors discovered in the pretrained LCM LoRA to an unmodified LDM (specifically, Stable Diffusion v1.5). As shown in Figure \ref{fig:ldm-transfer}, semantic offsets such as ``male → female'' successfully produce corresponding changes in the LDM outputs, indicating that the hidden-state ``H-spaces'' in the LCM and LDM are well-aligned in practice.

We acknowledge that our work does not exhaustively analyze how the LDM's latent representations evolve at each diffusion step. However, our primary objective is to understand how biases manifest in the final generated image, as this is where potential social and ethical impacts (e.g., reinforcement of harmful stereotypes) are most significant. While individual timesteps may encode different aspects of semantic information, it is ultimately the model's final output that determines how such biases become visible and impactful in practice.

\section{Normalization Methods for Ranking Cosine Distances}

In Section \ref{sec:one-to-many-methods}, we proposed a method for ranking descriptors based on their relative distances to target concepts. Here, we provide additional details on alternative normalization techniques and explain why mean-centering was selected as the primary method.

\subsection{Mean-Centering Normalization}

The primary approach we use for ranking is mean-centering, which adjusts the cosine distances by subtracting the mean distance across all target concepts:
\begin{equation}
    \begin{aligned}
        d'_{i,j} &= d_{i,j} - \frac{\sum_{k \in [n]}{d_{i,k}}}{n} \\
        &= d_{i,j} - \mu_j
    \end{aligned}
\end{equation}
where $d_{i,j}$ represents the cosine distance between descriptor $i$ and target concept $j$, and $n$ is the number of target concepts. $\mu_j$ is the mean of distances for the target concept $j$. This normalization ensures that rankings reflect relative similarity rather than absolute distance in the latent space.

\textbf{Justification:} Mean-centering effectively corrects for global shifts in distance while preserving relative rankings. This ensures that concepts closer to one target remain properly ranked even if their absolute distances vary. It is also very simple to calculate.

\subsection{Standard Deviation Scaling}

Another approach normalizes distances by dividing by the standard deviation of distances across all target concepts:
\begin{equation}
    d'_{i,j} = \frac{d_{i,j} - \mu_j}{\sigma_j}
\end{equation}
where $\mu_j$ and $\sigma_j$ are the mean and standard deviation of distances for target concept $j$.

\textbf{Justification:} Standard deviation scaling accounts for varying spread among distances.



\subsection{Subspace Projection Normalization}

An alternative method involves projecting each descriptor into the subspace spanned by the target concepts. we utilize Principal Component Analysis (PCA) to identify the most significant directions within the target concept space. Given a set of target concept vectors $V = \{ v_1, v_2, \dots, v_n \}$, we perform PCA to compute the principal components that best capture the variance in these vectors. Let $W$ be the matrix formed by stacking the top $M$ principal components as row vectors. The descriptor vector $v_i$ is then projected onto the principal subspace as follows:
\begin{equation}
    P_{\text{PCA}}(v_i) = W v_i
\end{equation}
where $P_{\text{PCA}}(v_i)$ is the projected vector in the subspace spanned by the principal components. The normalized distance is then computed as:
\begin{equation}
    d'_{i,j} = \text{cosine distance} \big( P_{\text{PCA}}(v_j), P_{\text{PCA}}(v_i) \big)
\end{equation}

\begin{figure}
    \centering
    \subfloat[]{
    \includegraphics[width=.35\linewidth]{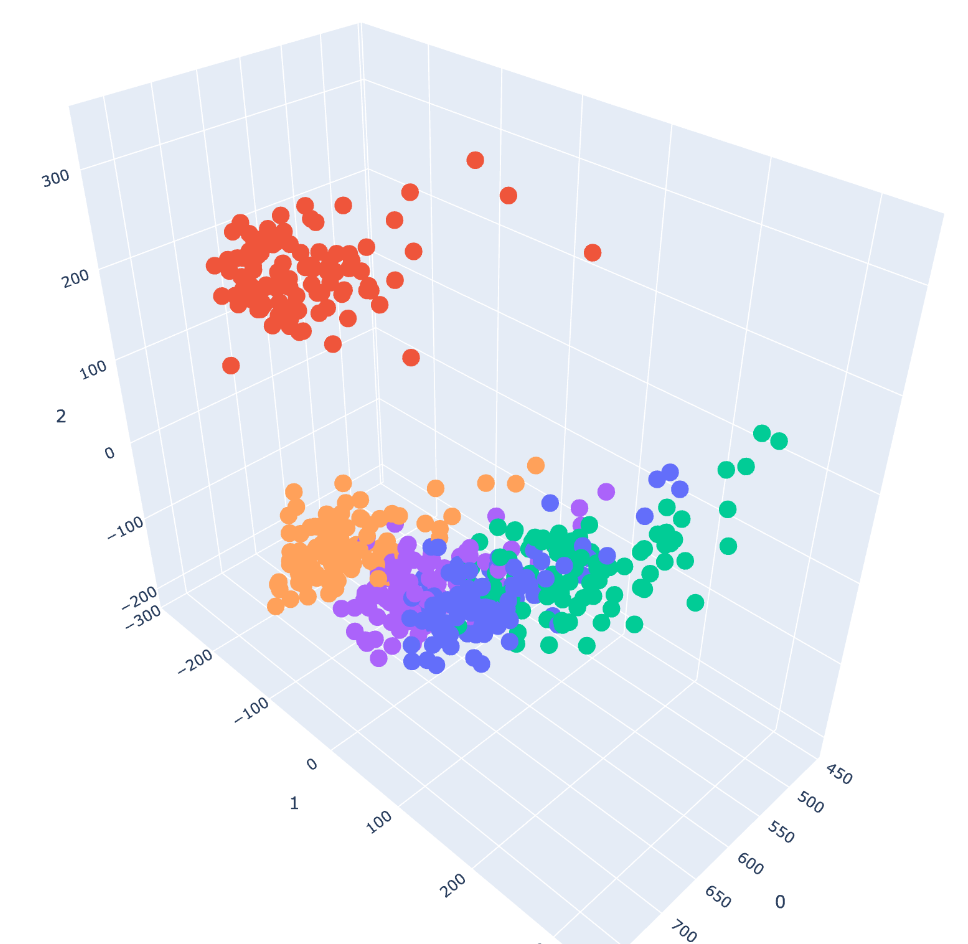}
    \label{fig:svd_race_uncentered}
    }
    \subfloat[]{
    \includegraphics[width=.55\linewidth]{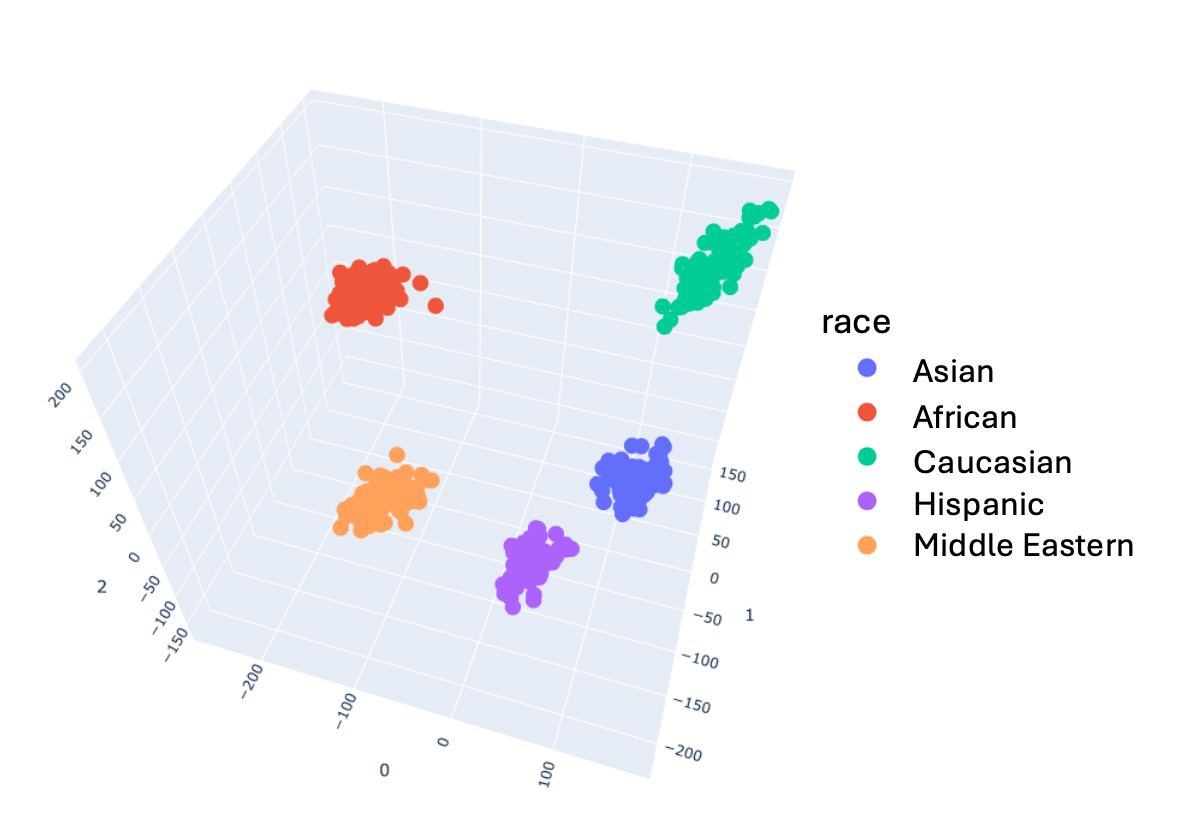}
    \label{fig:svd_race_centered}
    }
    \caption{Comparison of PCA projections before and after centering. (a) Without centering, projected vectors exhibit significant offsets and misalignment, making it harder to distinguish clusters. (b) After centering, vectors are more clearly separated, improving the interpretability of latent space organization.}
    \label{fig:svd_race}
\end{figure}

In practice, it is also important to center the target concept vectors at each seed. Figure \ref{fig:svd_race} highlights the impact of centering before PCA projection. Without centering (Figure \ref{fig:svd_race_uncentered}), latent vectors remain offset, leading to suboptimal separation of clusters. After centering (Figure \ref{fig:svd_race_centered}), the PCA projection aligns more cleanly, enhancing the interpretability of semantic distances.

\textbf{Justification:} PCA-based subspace projection helps remove components unrelated to the target concepts while retaining the most significant variations. This method ensures that rankings are based on the dominant semantic dimensions rather than noise or unrelated features. However, PCA assumes that the principal components correspond to meaningful semantic differences, which may not always hold. Additionally, if target concepts are highly entangled, PCA may not fully disentangle them, leading to potential misinterpretations.

\subsection{Empirical Comparison}

\begin{figure}
    \subfloat[all seeds]{
        \includegraphics[width=\linewidth]{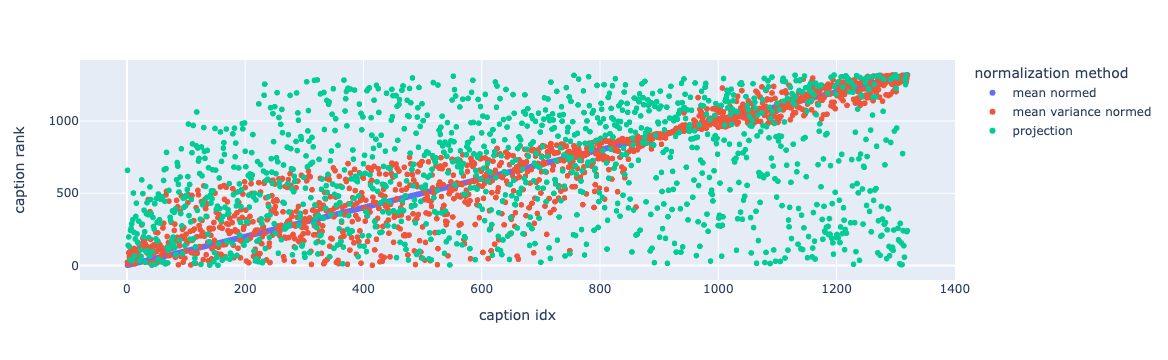}
        \label{fig:normalization_methods_rankings_all}
    }\\
    \subfloat[averaged on each seed]{
        \includegraphics[width=\linewidth]{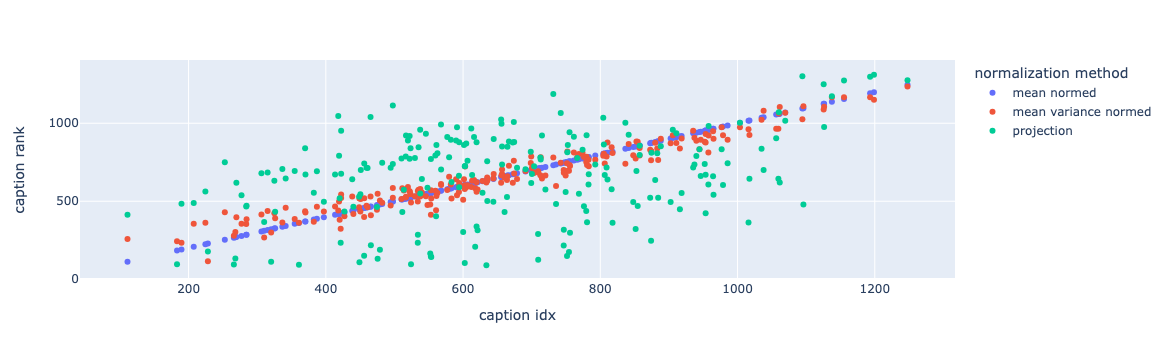}
        \label{fig:normalization_methods_rankings_avg}
    }
    
    \caption{Comparison of ranking results using different normalization techniques. Mean and variance normalization yield similar rankings, while PCA-based projection produces more distinct variations.}
    \label{fig:normalization_methods_rankings}
\end{figure}

Figure \ref{fig:normalization_methods_rankings} presents the rankings assigned to each caption using the three different normalization techniques. In Figure \ref{fig:normalization_methods_rankings_avg}, the ranking for each caption is averaged across the seeds, while figure \ref{fig:normalization_methods_rankings_all} plots every sample individually. The results indicate that normalizing by both variance and mean produces rankings similar to those obtained using mean normalization alone. In contrast, normalization through PCA-based projection often results in significantly different rankings, but the overall trend remains similar, especially at the extreme ends.

\textbf{Conclusion:} Mean-centering was selected as the primary normalization method due to its balance of robustness, interpretability, and computational efficiency. While subspace projection and standard deviation scaling offer alternative perspectives, they introduce additional assumptions that may not generalize well across datasets.

\section{Implementation Details}
\label{sec:implementation}

\paragraph{H-space extraction.}
For all experiments, H-space vectors were taken from the \texttt{output\_middle\_block}
of the U-Net in Stable Diffusion v1.5 and SDXL. The tensor shape at this layer is $(B, 1280, 8, 8)$ for SD1.5 and $(B, 1280, 32, 32)$ for SDXL. Cosine distances, clustering, and conditioning are all computed directly in this space, treating the flattened $(C \times H \times W)$ representation as the prompt-aligned
H-vector.

\paragraph{CLIP classification}
We rely on CLIP’s recommended zero-shot inference mode as a classifier, comparing \verb|similarity(image, "a photo of a man")| versus \verb|similarity(image, "a photo of a woman")| and taking the class of highest similarity. Comparing against 1800 manually labelled images generated during the experiments in this paper, CLIP has a classification accuracy of 96\%.

\paragraph{Computational efficiency.}
All experiments were conducted on a single NVIDIA RTX A6000 GPU (48GB). Extraction
requires $\sim$1.1s per prompt on SD1.5 and $\sim$6.3s per prompt on SDXL. Importantly, SCALEX requires no
training and scales linearly with the number of prompts.

\section{Prompt variations}
\label{sec:prompt_variations}

\begin{figure}[t]
  \centering

  \begin{subfigure}[t]{0.48\linewidth}
    \centering
    \includegraphics[width=\linewidth]{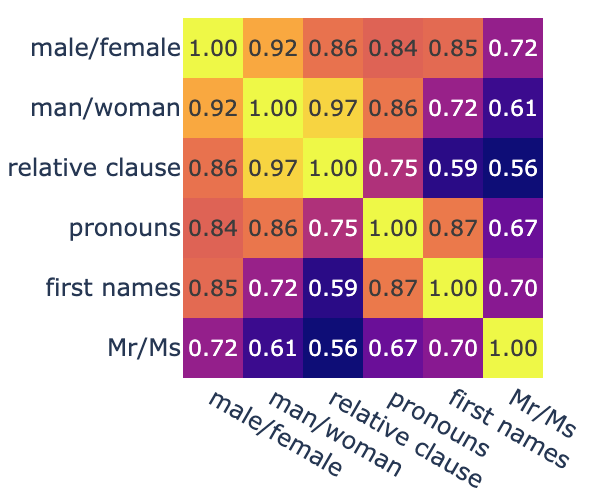}
    \caption{Pearson correlation with baseline}
    \label{fig:sd1_5_pearson}
  \end{subfigure}
  \hfill
  \begin{subfigure}[t]{0.48\linewidth}
    \centering
    \includegraphics[width=\linewidth]{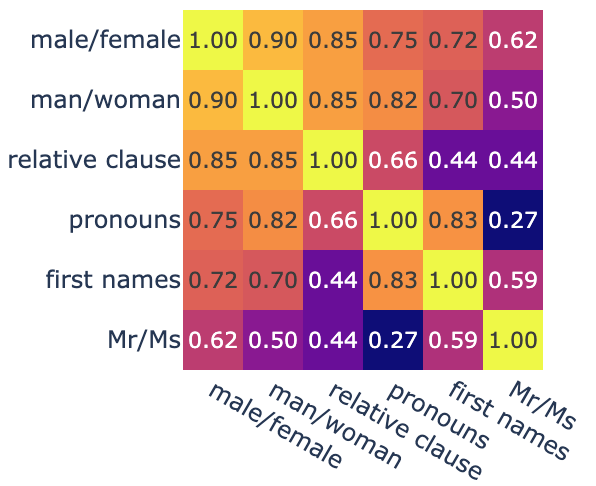}
    \caption{Spearman correlation with baseline}
    \label{fig:sd1_5_spearman}
  \end{subfigure}

  \caption{Robustness of one-to-one gender bias measurements across synonymous prompt variations in Stable Diffusion~1.5. Both (a) Pearson and (b) Spearman correlations are computed between the baseline \emph{female/male} phrasing and five alternatives: \emph{man/woman}, relative clause (\emph{a doctor who is a woman/man}), pronouns (\emph{she/he is a doctor}), first names (\emph{Sarah/John, a doctor}), and honorifics (\emph{Ms./Mr.\ surname}). Results confirm strong alignment for \emph{man/woman}, relative clause, and pronouns. First names and honorifics remain positively correlated but weaker, reflecting additional priors (e.g., age, cultural context in names) and subtler gender cues.}
  \label{fig:prompt_variation_corr}
\end{figure}

To evaluate the robustness of SCALEX against synonymous phrasings of gender markers,
we repeated the one-to-one comparison experiment described in Section~4.1 with five
common variations: \emph{man/woman}, relative clause
(\emph{a doctor who is a woman/man}), pronouns (\emph{she/he is a doctor}),
first names (\emph{Sarah/John, a doctor}), and honorifics
(\emph{Ms./Mr.\ surname}). Each profession prompt was paired with its male and female
variants, and cosine distance differences ($\Delta$ in Eq.~2) were computed relative to
the neutral form.

Figure~\ref{fig:prompt_variation_corr} reports both Pearson and Spearman correlations
with the baseline \emph{female/male} phrasing across all professions. Results show that
\emph{man/woman}, relative clause, and pronoun variants very similar bias
patterns (Pearson $r \in [0.84, 0.92]$), confirming that the one-to-one comparison is
stable to small wording changes. First-name variants (e.g., ``Sarah'' vs.~``John'') and
honorifics (``Ms.'' vs.~``Mr.'') also correlate positively ($r=0.85$ and $r=0.72$,
respectively), but with reduced strength. We attribute this to two factors:
(i) personal names often introduce additional implicit attributes such as age or cultural priors, and
(ii) honorifics are weaker gender cues in large-scale caption distributions.

Together, these findings validate that SCALEX captures stable gender bias signals
independent of synonym choice, while also revealing when additional attributes
become entangled with the gender markers. The weaker correlations for names and
honorifics highlight an opportunity for SCALEX to surface broader, contextualized
biases that extend beyond simple gender markers.

\section{SDXL results}
\label{sec:sdxl}

We repeat the synonym-robustness experiments from Section~\ref{sec:prompt_variations}
on Stable Diffusion XL (SDXL). As with SD~1.5, we compare five common variants of
gender markers (\emph{man/woman}, relative clause, pronouns, first names, and honorifics)
against the baseline \emph{female/male} phrasing across all professions.

\begin{figure}[t]
    \centering
    \includegraphics[width=0.75\linewidth]{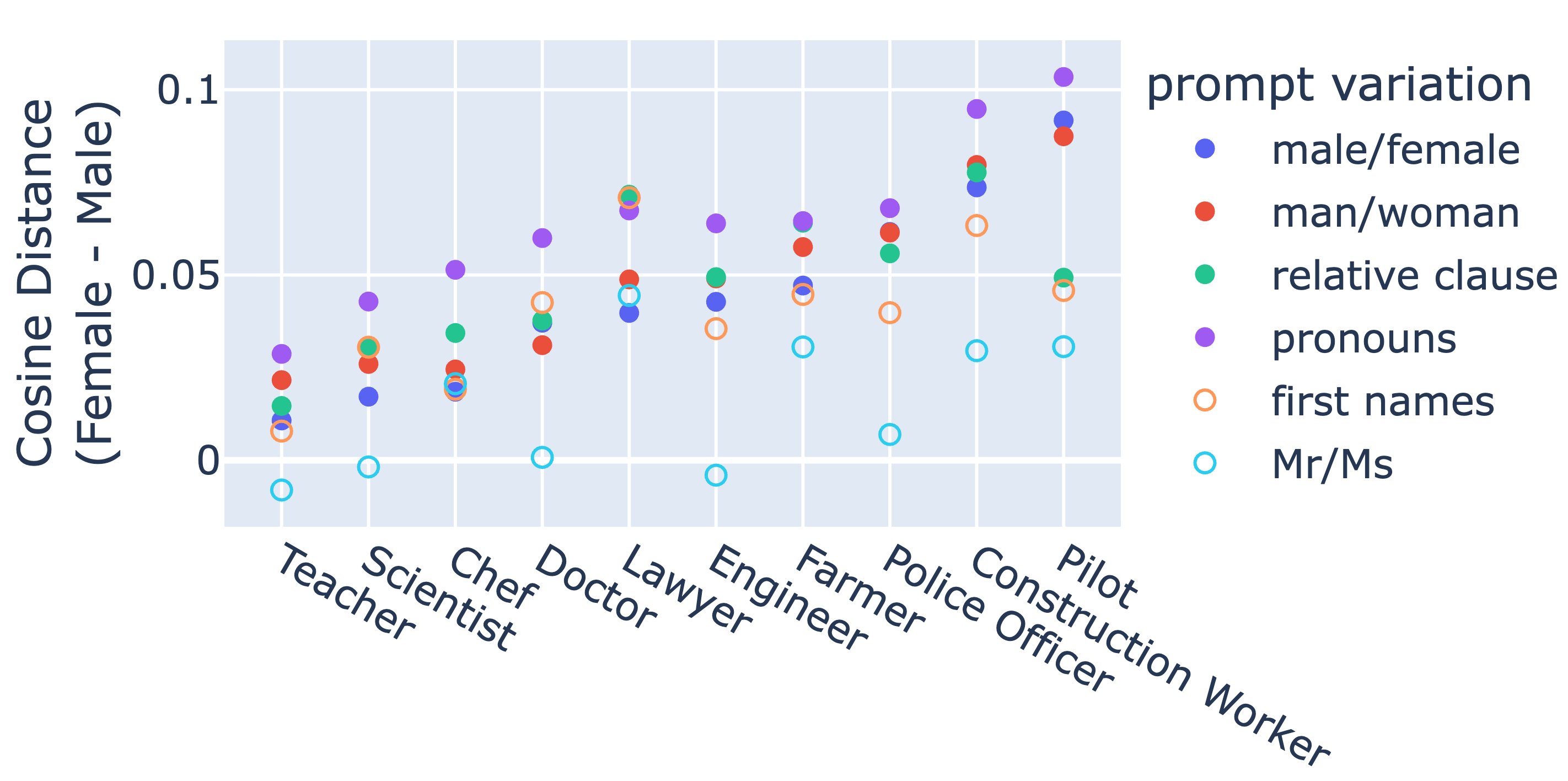}
    \caption{Average cosine distance differences $\Delta$ (Eq.~2) between female and male
    prompts across professions in SDXL. Patterns are consistent with SD~1.5, showing
    strong defaults toward male-aligned representations in professions such as pilot,
    construction worker, and police officer, while teacher remains relatively balanced.
    Notably, the overall magnitude of cosine distances is higher in SDXL, aligning with
    Table~\ref{tab:professions_percent_female} where SDXL produced a lower percentage of
    female-presenting images. Together, these results suggest that SDXL exhibits
    \emph{stronger gender bias} than SD~1.5 despite producing higher-quality images.}
    \label{fig:sdxl_profession_bias}
\end{figure}

\begin{figure}[t]
  \centering
  \begin{subfigure}[t]{0.48\linewidth}
    \centering
    \includegraphics[width=\linewidth]{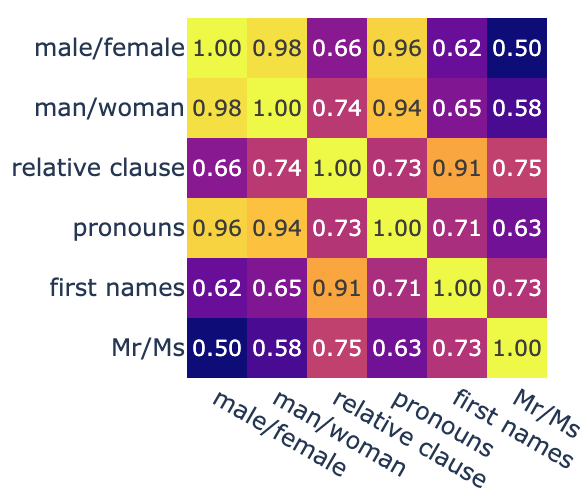}
    \caption{Pearson correlation with baseline}
    \label{fig:sdxl_pearson}
  \end{subfigure}
  \hfill
  \begin{subfigure}[t]{0.48\linewidth}
    \centering
    \includegraphics[width=\linewidth]{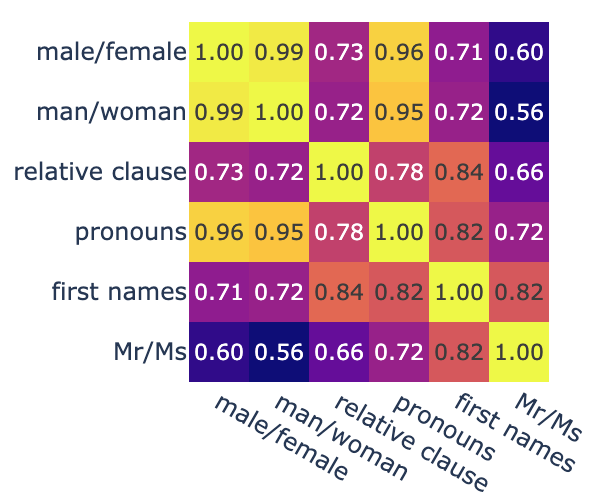}
    \caption{Spearman correlation with baseline}
    \label{fig:sdxl_spearman}
  \end{subfigure}
  \caption{Robustness of one-to-one gender bias measurements across synonymous
  prompt variations in SDXL. Both Pearson and Spearman correlations confirm strong
  alignment for \emph{man/woman} and pronouns ($r \geq 0.96$). However, SDXL shows
  greater sensitivity to the relative clause variant ($r = 0.66$), and first names
  and honorifics correlate even more weakly than in SD~1.5 ($r = 0.62$ and $r = 0.50$).
  These results indicate that, while SDXL achieves visually higher-quality generations,
  it is also more sensitive to prompt wording.}
  \label{fig:sdxl_prompt_variation_corr}
\end{figure}

Across both architectures, SCALEX captures consistent bias patterns but also reveals
systematic differences: SD~1.5 shows greater robustness to phrasing, while SDXL
displays stronger overall gender bias and higher sensitivity to linguistic variation.

\section{Prompts used}

We provide the list of prompts generated using ChatGPT \cite{openai2024chatgpt} for defaults and descriptor experiments

\subsection{Professions}
\label{sec:professions_prompts}
Used for one-to-one comparisons in Section \ref{sec:one-to-one-experiment}

\begin{enumerate}
    \item A doctor in a white coat, consulting with a patient in a hospital room.
    \item A doctor in a lab coat, reviewing medical charts in a modern hospital.
    \item A doctor examining an X-ray in a radiology room.
    \item A doctor performing surgery in an operating room.
    \item A doctor speaking with a nurse in a busy emergency room.
    \item A doctor providing a check-up to a child in a clinic.
    \item A doctor performing a diagnostic procedure with medical equipment.
    \item A doctor discussing patient care with a team in a hospital hallway.
    \item A doctor giving a presentation during a medical conference.
    \item A doctor in a white coat, standing by an ultrasound machine in a maternity ward.
    \item An engineer working on machinery in a factory setting.
    \item An engineer inspecting a blueprint at a construction site.
    \item An engineer programming a robotic arm in a high-tech lab.
    \item An engineer supervising an assembly line in a manufacturing plant.
    \item An engineer working on a bridge construction project.
    \item An engineer running diagnostics on electronic circuits in an engineering lab.
    \item An engineer working on renewable energy systems at a wind farm.
    \item An engineer designing a new product on CAD software in an office.
    \item An engineer presenting plans in an architectural firm’s boardroom.
    \item An engineer wearing a hard hat, inspecting materials on a job site.
    \item A teacher instructing students in a high school science class.
    \item A teacher reading a book aloud to children in an elementary classroom.
    \item A teacher writing equations on a chalkboard in a math class.
    \item A teacher leading a discussion in a university lecture hall.
    \item A teacher helping a student with a chemistry experiment in a lab.
    \item A teacher guiding students in a group project in a classroom.
    \item A teacher teaching a history class using interactive technology.
    \item A teacher organizing a field trip with her students outside.
    \item A teacher instructing students during a physical education class.
    \item A teacher leading an art class in a creative studio.
    \item A lawyer presenting a case in a courtroom in front of a jury.
    \item A lawyer consulting with clients in a modern law firm.
    \item A lawyer reviewing documents at his desk before a trial.
    \item A lawyer making an argument in front of a judge in court.
    \item A lawyer negotiating a settlement with opposing counsel in a conference room.
    \item A lawyer speaking with a client in a prison meeting room.
    \item A lawyer preparing for a case in a courtroom hallway.
    \item A lawyer researching case law in a legal library.
    \item A lawyer attending a mediation session in a law office.
    \item A lawyer participating in a corporate board meeting as legal counsel.
    \item A chef plating a gourmet dish in a high-end restaurant kitchen.
    \item A chef preparing sushi in a busy kitchen environment.
    \item A chef chopping vegetables in a fast-paced kitchen.
    \item A chef baking pastries in a bakery’s kitchen.
    \item A chef grilling food in an outdoor kitchen for an event.
    \item A chef instructing culinary students in a professional kitchen.
    \item A chef working in a seafood restaurant, preparing fresh fish.
    \item A chef arranging desserts in a Michelin-starred restaurant.
    \item A chef preparing traditional dishes in a local restaurant.
    \item A chef leading a team in a catering kitchen for a large event.
    \item A scientist conducting research in a sterile laboratory.
    \item A scientist analyzing data on a computer in a tech lab.
    \item A scientist performing experiments with chemicals in a lab.
    \item A scientist using a microscope in a medical research lab.
    \item A scientist explaining his findings at a scientific conference.
    \item A scientist working with biological samples in a medical facility.
    \item A scientist in a clean room, working with nanotechnology.
    \item A scientist presenting her research findings to colleagues.
    \item A scientist inspecting samples in an environmental research lab.
    \item A scientist working with robotics in an advanced tech facility.
    \item A police officer patrolling a busy urban street.
    \item A police officer interviewing a witness at a crime scene.
    \item A police officer directing traffic in a bustling city center.
    \item A police officer conducting an arrest in a suburban neighborhood.
    \item A police officer giving a safety talk to a group of children.
    \item A police officer attending a briefing at the police station.
    \item A police officer investigating a burglary in a small town.
    \item A police officer examining evidence in a crime lab.
    \item A police officer patrolling a park on foot.
    \item A police officer performing a routine traffic stop on a highway.
    \item A construction worker operating a bulldozer at a construction site.
    \item A construction worker wearing a hard hat, inspecting building materials.
    \item A construction worker working on scaffolding at a skyscraper project.
    \item A construction worker laying bricks on a building project.
    \item A construction worker pouring cement for a foundation.
    \item A construction worker operating a crane at a construction site.
    \item A construction worker welding metal beams on a high-rise structure.
    \item A construction worker installing plumbing in a new building.
    \item A construction worker working on road repairs with heavy equipment.
    \item A construction worker managing a construction crew at a building site.
    \item A pilot sitting in the cockpit, preparing for takeoff.
    \item A pilot giving a pre-flight briefing to passengers.
    \item A pilot inspecting the aircraft before departure.
    \item A pilot controlling the plane during a smooth flight.
    \item A pilot wearing a uniform, standing next to a jet on the tarmac.
    \item A pilot reviewing flight plans in the control room.
    \item A pilot landing a commercial airplane at a busy airport.
    \item A pilot participating in a safety drill with the crew.
    \item A pilot flying a small private plane over a scenic landscape.
    \item A pilot handling air traffic communications from the cockpit.
    \item A farmer driving a tractor through a field.
    \item A farmer harvesting crops in a rural landscape.
    \item A farmer tending to cattle on a ranch.
    \item A farmer watering plants in a greenhouse.
    \item A farmer picking vegetables in a family-owned farm.
    \item A farmer supervising workers in a vineyard.
    \item A farmer loading harvested grain into a truck.
    \item A farmer checking soil quality on an organic farm.
    \item A farmer herding sheep across a pasture.
    \item A farmer selling fresh produce at a local farmer’s market.
\end{enumerate}

\subsection{Facial Features}

Used for one-to-many comparisons in Section \ref{sec:one-to-many-experiment}

\subsubsection{Facial Structure}

\begin{enumerate}
    \item Round face with soft, full cheeks and a gently curved jawline, accentuated by a short chin.
    \item Square face with a prominent, angular jawline and a broad forehead, creating a sharp and defined look.
    \item Heart-shaped face with a wide forehead and high cheekbones, tapering down to a pointed chin.
    \item Oval face with balanced proportions, featuring a slightly rounded jawline and a forehead that mirrors the width of the cheekbones.
    \item Long, rectangular face with a narrow chin and a high forehead, giving an elongated appearance.
    \item Diamond-shaped face with narrow forehead and jawline, and prominent, wide cheekbones.
    \item Triangular face with a broad, flat jawline and a narrower forehead, creating a strong, tapered structure.
    \item Oblong face with a balanced but longer vertical shape, featuring a rounded chin and a forehead slightly wider than the jawline.
    \item Softly square face with a defined jawline that is less sharp, paired with a broad forehead and cheeks.
    \item Petite, heart-shaped face with delicate, high cheekbones and a narrow, pointed chin.
    \item Wide, round face with full cheeks and a smooth jawline that curves gently toward the chin.
    \item Chiseled, angular face with a strong jawline, high cheekbones, and a pronounced, narrow chin.
    \item Broad, square face with a defined jawline, prominent chin, and a wide, flat forehead.
    \item Soft, oval face with a slightly tapered jawline and rounded forehead, giving a balanced and gentle appearance.
    \item Diamond-shaped face with angular features, including a narrow forehead, wide cheekbones, and a sharp chin.
    \item Elongated face with a long, narrow jawline and a high forehead, giving a sleek, vertical appearance.
    \item Compact, round face with a small chin, soft cheeks, and a smooth, rounded hairline.
    \item Defined, heart-shaped face with high, pronounced cheekbones and a small, pointed chin.
    \item Angular, square face with a broad jawline and a flat, strong forehead, creating a powerful and bold structure.
    \item Narrow, oblong face with soft edges, featuring a gently pointed chin and high, rounded cheekbones.
\end{enumerate}

\subsubsection{Hair}

\begin{enumerate}
    \item Short, neatly trimmed hair with a side part and slight wave.
    \item Long, straight hair cascading down the shoulders, with bangs framing the forehead.
    \item Curly, shoulder-length hair with tight ringlets, slightly frizzy at the ends.
    \item Buzz cut with a clean-shaven look, highlighting a well-defined jawline.
    \item Medium-length, wavy hair pulled back into a loose ponytail with soft tendrils.
    \item Slicked-back, gelled hair with a sharp undercut on the sides.
    \item Short, spiky hair with frosted tips and a textured finish.
    \item Long, braided hair, styled into a single thick braid that falls over the shoulder.
    \item Messy bob with tousled waves and an asymmetrical cut that grazes the chin.
    \item Bald head with a light sheen, complemented by a well-groomed beard.
    \item Shoulder-length hair with soft curls and subtle highlights throughout.
    \item Short, pixie cut with choppy layers and a fringe covering one eye.
    \item Sleek, straight hair pulled back into a high bun, showing off a smooth hairline.
    \item Long, thick dreadlocks tied up into a half-up, half-down style.
    \item Curly afro, with tightly coiled hair forming a rounded shape.
    \item Shoulder-length hair in loose waves, casually tucked behind the ears.
    \item Short, cropped hair with a fade on the sides and a defined part.
    \item Medium-length hair in a messy topknot with loose strands framing the face.
    \item Straight, chin-length bob with a sharp, blunt cut and no layers.
    \item Long, wavy hair with ombre coloring that fades from dark brown to light blonde.
\end{enumerate}

\subsubsection{Eyes}

 \begin{enumerate}
     \item Large almond-shaped eyes, with a soft, thoughtful gaze.
     \item Narrow, hooded eyes, glinting with quiet determination.
     \item Round, wide-set eyes, sparkling with excitement.
     \item Small, deep-set eyes, casting a calm, serious expression.
     \item Bright, close-set eyes, crinkled at the corners from a wide smile.
     \item Piercing, upturned eyes, focused and intense.
     \item Soft, sleepy eyes, slightly drooping with a peaceful expression.
     \item Dark, downturned eyes, reflecting a sense of calm and contemplation.
     \item Wide, curious eyes, open and full of wonder.
     \item Sharp, cat-like eyes, with a playful gleam.
     \item Warm, kind eyes, with a gentle upward tilt at the edges.
     \item Deep-set eyes, shadowed by thick lashes, holding an air of mystery.
     \item Bright, round eyes, slightly squinting in laughter.
     \item Narrow, almond-shaped eyes, with an intense, focused look.
     \item Tired, half-lidded eyes, softened by a quiet smile.
     \item Clear, bright eyes, set wide apart, giving an innocent expression.
     \item Soft, downcast eyes, conveying quiet thoughtfulness.
     \item Bright, sharp eyes, alert and darting, full of energy.
     \item Deep-set, intense eyes, framed by dark, expressive eyebrows.
     \item Wide, sparkling eyes, brimming with curiosity and joy.
 \end{enumerate}

\subsubsection{Eyebrows}

\begin{enumerate}
    \item Sharp, arched eyebrows give a dramatic and intense look.
    \item Soft, straight eyebrows create a calm and neutral expression.
    \item Thick, bold eyebrows dominate the face, adding a strong, confident presence.
    \item Delicate, thin eyebrows that arch slightly, providing a gentle appearance.
    \item Bushy eyebrows with a natural, untamed look that conveys individuality.
    \item Perfectly groomed, high arches for a sleek and polished vibe.
    \item Rounded eyebrows, subtly shaping the face for a soft and approachable expression.
    \item Sparse eyebrows with a light, barely-there effect.
    \item Thick, straight eyebrows that sit low on the brow bone, giving an intense gaze.
    \item Sharp-angled eyebrows that rise dramatically, emphasizing surprise or alertness.
    \item Flat, wide eyebrows that give a bold and direct appearance.
    \item Thin, high-arched eyebrows add a vintage, classic flair to the face.
    \item Naturally thick eyebrows with slight curves, offering a relaxed, carefree look.
    \item Short, straight eyebrows close to the eyes, creating a focused and determined expression.
    \item Softly curved eyebrows that subtly lift at the outer corners, giving a hint of curiosity.
    \item Thick, upward-sweeping brows that provide a bold and energetic vibe.
    \item Barely noticeable, faint eyebrows create a gentle and understated look.
    \item Well-defined, symmetrical eyebrows frame the face with precision and balance.
    \item Thin, gently sloping eyebrows that add a wistful, dreamy quality to the face.
    \item Naturally thick, slightly uneven eyebrows give a quirky, playful character.
\end{enumerate}

\subsubsection{Nose}

\begin{enumerate}
    \item A petite button nose with a slight upward tilt, giving a youthful appearance.
    \item A broad, flat nose with a gentle slope down the center.
    \item A long, straight nose with a sharp bridge and narrow nostrils.
    \item A rounded nose with a bulbous tip, creating a soft and friendly look.
    \item A delicate, narrow nose with a smooth, defined bridge.
    \item A wide nose with flared nostrils and a flat bridge.
    \item A slightly crooked nose with a noticeable bump on the bridge.
    \item A prominent aquiline nose with a strong downward curve at the tip.
    \item A short, upturned nose with a subtle ridge.
    \item A straight, narrow nose with a high bridge and a defined tip.
    \item A large, rounded nose with wide nostrils, giving the face a bold presence.
    \item A small, slender nose with a gentle slope and narrow nostrils.
    \item A wide, flat nose with a broad bridge and rounded tip.
    \item A long, pointed nose with sharp angles and a slightly hooked tip.
    \item A short, wide nose with a flat bridge and round nostrils.
    \item A thin, angular nose with a sharp, defined bridge and pointed tip.
    \item A prominent Roman nose with a curved bridge and pointed tip.
    \item A small, delicate nose with a slight upturn and narrow nostrils.
    \item A broad, low-bridge nose with a rounded, prominent tip.
    \item A high-bridged nose with a strong, straight profile and defined nostrils.
\end{enumerate}

\subsubsection{Mouth}

\begin{enumerate}
    \item A broad smile reveals a set of perfectly straight teeth, lips slightly parted.
    \item Thin lips, tightly pressed, creating a serious and contemplative expression.
    \item Full, glossy lips form a slight smirk, adding a playful touch to the face.
    \item The corners of the lips turn upward in a gentle, serene smile, with dimples showing.
    \item Wide, expressive mouth with slightly downturned lips, reflecting a thoughtful expression.
    \item A subtle grin with full lips that creates a warm and inviting look.
    \item Pursed lips, as if mid-thought, with a hint of tension around the edges.
    \item A toothy smile, with lips stretched wide, radiating joy and excitement.
    \item Slightly chapped lips are pulled into a neutral line, giving a calm, relaxed vibe.
    \item Small, narrow lips slightly curved, giving a quiet, gentle demeanor.
    \item Full lips, slightly open, revealing a soft, natural pout in a relaxed expression.
    \item The upper lip is thinner than the lower, both pulled into a soft, closed-mouth smile.
    \item Lips are softly parted as if mid-sentence, with a natural, unpolished feel.
    \item A playful grin with lips pursed, as if holding back laughter.
    \item A deep frown, with lips turned down sharply at the corners, adding intensity to the face.
    \item Wide, thin lips pulled into a broad, confident smile, showing just a few teeth.
    \item A shy smile with lips closed tightly, but the corners lift slightly in a sweet expression.
    \item The lips form a soft, natural curve, slightly upturned, giving a peaceful expression.
    \item Full lips slightly pursed in concentration, the upper lip pronounced and defined.
    \item Lips are relaxed, slightly parted with a faint hint of a smile, showing calmness.
\end{enumerate}

\subsubsection{Skin features}

\begin{enumerate}
    \item Smooth complexion with faint freckles scattered across the nose and cheeks.
    \item Prominent laugh lines around the mouth, with a small mole near the right eyebrow.
    \item Sun-kissed skin with a subtle sheen and a light scar above the left cheekbone.
    \item Youthful skin with an even tone, accented by a cluster of faint freckles on the forehead.
    \item Slightly weathered skin with visible crow's feet around the eyes and a small birthmark near the chin.
    \item Clear complexion with a single, dark mole on the left cheek and faint acne scars on the forehead.
    \item Deep smile lines around the mouth, with a subtle sunspot on the upper cheek.
    \item Smooth skin with a pale tone and a few small, raised freckles across the bridge of the nose.
    \item Tanned skin with visible sunspots and a faint scar running along the right jawline.
    \item Porcelain-like skin with a light dusting of freckles on the cheeks and a tiny scar near the hairline.
    \item Clear, even-toned skin with pronounced forehead lines and a single dimple on the right cheek.
    \item Mature skin with visible age spots and prominent frown lines between the eyebrows.
    \item Slightly dry skin with a patch of freckles along the left cheek and faint redness around the nose.
    \item Smooth, glowing skin with a soft sheen and a single mole just above the upper lip.
    \item Clear complexion with a small acne scar on the chin and subtle smile lines around the eyes.
    \item Fair skin with a light tan, dotted with a few prominent freckles along the upper cheekbones.
    \item Smooth skin with a rosy hue, a small birthmark near the left eye, and a faint scar under the chin.
    \item Even skin tone with slight redness around the nose and small, visible pores on the cheeks.
    \item Youthful skin with a clear complexion, a light scar on the forehead, and faint freckles around the mouth.
    \item Soft skin with a slight sheen, visible laugh lines, and a single mole on the left temple.
\end{enumerate}

\subsubsection{Facial Hair}

 \begin{enumerate}
     \item Thick, full beard with a neatly trimmed mustache, framing a strong jawline.
     \item A subtle shadow of stubble along the jaw, giving a rugged, yet clean look.
     \item Clean-shaven face with a sharp, defined goatee around the chin.
     \item Long, flowing beard with a wild, unkempt look, matching a bushy mustache.
     \item Smoothly shaven face with only a faint mustache, barely noticeable.
     \item A handlebar mustache curled upward, paired with a small, pointed beard on the chin.
     \item Short boxed beard with a neatly sculpted outline, emphasizing the cheeks.
     \item Bare face except for a pencil-thin mustache that adds a touch of elegance.
     \item A thick, bushy beard covering the lower half of the face, with a trimmed mustache.
     \item Chinstrap beard extending from ear to ear, perfectly framing the jaw.
     \item Patchy stubble with a scruffy appearance, contrasting with a clean-shaven upper lip.
     \item A neatly shaped Van Dyke beard with a sharp mustache, creating a distinguished look.
     \item Classic soul patch below the lower lip, with the rest of the face clean-shaven.
     \item Mutton chop sideburns that connect with a trimmed goatee for a bold statement.
     \item Short, even stubble all over the face, adding texture and definition.
     \item Thick mustache paired with a trimmed beard that follows the contour of the chin.
     \item Full beard with natural curls, giving a soft and voluminous appearance.
     \item Bare face except for a small, neatly trimmed goatee, adding subtle definition.
     \item Clean-shaven except for a rugged stubble along the jawline and chin.
     \item A long, tapered beard with a sleek mustache, creating a regal and polished style."
 \end{enumerate}

\subsubsection{Forehead}

\begin{enumerate}
    \item A high, smooth forehead with a subtle widow’s peak adds elegance to the overall facial structure.
    \item A narrow forehead with deep-set expression lines that emphasize a thoughtful demeanor.
    \item The broad forehead is slightly wrinkled, giving a wise and contemplative appearance.
    \item A low, rounded forehead with a smooth surface, complementing soft facial features.
    \item A prominent forehead with a slight protrusion and faint frown lines across the top.
    \item The angular forehead slopes gently, with a few horizontal lines adding character.
    \item A wide, smooth forehead with a visible hairline curve, balancing sharp cheekbones.
    \item A narrow, high forehead with a few vertical lines between the brows, giving a focused expression.
    \item A flat, even forehead that enhances the symmetrical proportions of the face.
    \item A gently sloping forehead with faint, natural wrinkles just above the eyebrows.
    \item A wide forehead framed by wisps of hair, with a noticeable cleft in the middle of the brow.
    \item The smooth, reflective forehead contrasts with deep-set eyes, drawing attention upward.
    \item A small, low-set forehead with fine creases, giving a serene and calm look.
    \item A broad forehead with deep horizontal wrinkles, emphasizing a mature and experienced face.
    \item The narrow forehead is smooth and seamless, making the face look youthful and fresh.
    \item A rounded, high forehead with subtle lines that soften the sharp jawline.
    \item A slightly wrinkled forehead with a pronounced V-shaped hairline, hinting at concentration.
    \item The sloping forehead has a few fine lines, showing age gracefully while maintaining balance.
    \item A smooth, glossy forehead with no visible lines, giving a vibrant and animated expression.
    \item The wide forehead has a prominent furrow just above the brows, adding intensity to the gaze.
\end{enumerate}

\subsubsection{Chin}

\begin{enumerate}
    \item A sharp, pointed chin with a subtle dimple adds definition to the face.
    \item A rounded chin that softens the strong jawline, giving a youthful appearance.
    \item A prominent, square chin that creates a bold, angular profile.
    \item A small, delicate chin that tapers gently toward the neck.
    \item A chin with a deep cleft, adding character and distinctiveness to the face.
    \item A smooth, rounded chin that balances the facial features with gentle curves.
    \item A wide, strong chin that enhances the masculine structure of the face.
    \item A petite chin with a slight upward tilt, giving the face a playful expression.
    \item A narrow, elongated chin that contributes to an elegant, refined look.
    \item A softly rounded chin with a faint line running through the center.
    \item A bold, jutting chin that gives the face a determined, confident look.
    \item A delicate, pointed chin that contrasts with fuller cheeks, creating balance.
    \item A well-defined, square chin with subtle shadows along the jawline.
    \item A gently sloped chin that blends smoothly with the neck, adding gracefulness.
    \item A firm, angular chin with a slight indentation in the center, highlighting symmetry.
    \item A rounded chin with a prominent cleft, adding a unique touch to the face.
    \item A small, subtle chin that enhances the softness of the overall facial features.
    \item A broad chin with sharp lines that emphasize the face's strong geometric angles.
    \item A chin with a slight dimple, giving a charming, approachable look.
    \item A chin that softly curves inward, creating a gentle, feminine silhouette.
\end{enumerate}

\subsubsection{Expression}

\begin{enumerate}
    \item A face glowing with a warm, welcoming smile, eyes crinkling with genuine joy.
    \item A serious, contemplative look, lips pressed together and brow slightly furrowed.
    \item A face radiating surprise, mouth slightly open and eyes wide in amazement.
    \item A soft, relaxed expression, the gaze peaceful and content.
    \item A mischievous grin, eyes twinkling with playful energy.
    \item A face showing deep concentration, lips pursed and eyebrows drawn in.
    \item A calm and serene expression, eyes closed and a gentle smile playing on the lips.
    \item An intense, focused stare, eyes sharp and unwavering.
    \item A face beaming with excitement, cheeks flushed and smile broad.
    \item A blank, neutral expression, gaze steady and unreadable.
    \item A face marked by sorrow, eyes watery and lips turned downward in sadness.
    \item An amused expression, one eyebrow raised and a slight smirk on the lips.
    \item A face displaying confusion, eyebrows knit together and mouth slightly ajar.
    \item A look of quiet determination, lips set firmly and eyes focused ahead.
    \item A joyful expression with laughter in the eyes, mouth open in mid-laugh.
    \item A face showing shock, eyes wide and jaw slightly dropped in disbelief.
    \item A look of deep empathy, eyes soft and lips gently curved in a comforting smile.
    \item A face full of curiosity, head tilted slightly, eyes bright with wonder.
    \item A look of disappointment, lips pursed and eyes gazing downward.
    \item A proud expression, chin slightly lifted, with a subtle smile of accomplishment.
\end{enumerate}

\subsubsection{Ears}

\begin{enumerate}
    \item Small, close-set ears tucked neatly against the head.
    \item Large, slightly protruding ears with a rounded upper edge.
    \item Medium-sized ears with a subtle point at the top, positioned symmetrically.
    \item Ears with detached lobes, featuring a slight inward curve at the middle.
    \item Compact ears with multiple piercings on the lobes and upper cartilage.
    \item Ears that stick out noticeably, giving the face a quirky and unique charm.
    \item Larger ears with wide, fleshy lobes, pierced with small silver hoops.
    \item Small, rounded ears with an attached lobe, closely aligned to the head.
    \item Ears with a sharp, angular top edge and a deep fold along the inner ridge.
    \item Long, narrow ears with subtle curves and a smooth contour along the edges.
    \item Ears with prominent upper cartilage, slightly folded forward at the tips.
    \item Ears with thick, rounded lobes and a soft, smooth texture.
    \item Wide ears with a prominent outer rim and a natural outward flare.
    \item Small ears with a slight asymmetry, the right ear sitting lower than the left.
    \item Larger ears with a pronounced tragus and several cartilage piercings.
    \item Medium-sized ears with an elegant taper at the top and no visible piercings.
    \item Ears with a sharp upper curve and a thick, round lobe adorned with a stud.
    \item Flat, close-to-head ears with a faint crease along the outer edge.
    \item Oval-shaped ears with elongated lobes, giving the face a soft, gentle appearance.
    \item Compact ears with an attached lobe, featuring a subtle diamond-shaped contour.
\end{enumerate}


\end{document}